\documentclass[lettersize,journal]{IEEEtran}
\usepackage{amsmath,amsfonts}
\usepackage{algorithmic}
\usepackage{algorithm}
\usepackage{array}
\usepackage[caption=false,font=normalsize,labelfont=sf,textfont=sf]{subfig}
\usepackage{textcomp}
\usepackage{stfloats}
\usepackage{url}
\usepackage{verbatim}
\usepackage{graphicx}
\usepackage{cite}
\usepackage{booktabs} 
\usepackage{enumitem}
\usepackage{graphicx}
\usepackage{hyperref}

\begin{document}

\title{Unlock the Potential of Large Language Models for Predictive Tabular Tasks in Data Science with Table-Specific Pretraining}

\author{
Yazheng Yang\IEEEauthorrefmark{1},
Yuqi Wang\IEEEauthorrefmark{1},
Yaxuan Li\IEEEauthorrefmark{2},
Sankalok Sen\IEEEauthorrefmark{1},
Lei Li\IEEEauthorrefmark{1},
Lin Qiu\IEEEauthorrefmark{3},
and Qi Liu\IEEEauthorrefmark{1}

\IEEEcompsocitemizethanks{

\IEEEcompsocthanksitem \IEEEauthorrefmark{1} Department of Computer Science,
The University of Hong Kong, Hong Kong, China. \IEEEauthorrefmark{2} School of Computer Science and Technology,
Harbin Institute of Technology, Shenzhen, China.
\IEEEauthorrefmark{3} Department of Information Systems and Management Engineering,
Southern University of Science and Technology, Shenzhen, China.
}
}



\maketitle

\begin{abstract}
In data science, predictive tasks such as classification, regression, and missing value imputation are fundamental challenges in tabular data analysis. This research investigates the application of Large Language Models (LLMs) to these tasks. While LLMs excel in natural language understanding, their effectiveness on structured tabular data remains limited due to minimal exposure during pretraining. To address this gap, we construct a large-scale corpus of annotated tables and introduce a tailored pretraining framework. Our trained model achieves significant improvements over baselines, with an average gain of 8.9\% in classification and 10.7\% in regression tasks. We further evaluate its performance in zero-shot and few-shot prediction, as well as in-context learning scenarios. Extensive experiments demonstrate substantial gains over existing benchmarks, highlighting the potential of LLMs for tabular data processing. Additionally, we apply our approach across multiple open-source LLMs and demonstrate its generalizability. This work establishes a new benchmark for enhancing tabular intelligence through LLM-based pretraining.\footnote{Our trained model is publicly available.}
\end{abstract}


\section{Introduction}
\label{intro_sect}
\IEEEPARstart{T}{ables} are widely used across domains such as finance, data analytics, and logistics~\cite{nararatwong-etal-2022-enhancing,he-etal-2023-anameta,zhao-etal-2023-qtsumm,zhang-etal-2023-generative}. In data science, key tasks on tabular data include classification, regression, and missing-value imputation, all of which are fundamental to predictive modeling. Recent research increasingly focuses on leveraging advanced AI techniques to address these tasks effectively~\cite{gong2020tablegpt,zhang2023tablellama,zhao2023investigating,zhu2023xtab,slack2023tablet}. 

\begin{figure}[ht]
  \vskip 2ex
  \centering
  \includegraphics[width=1.0\columnwidth]{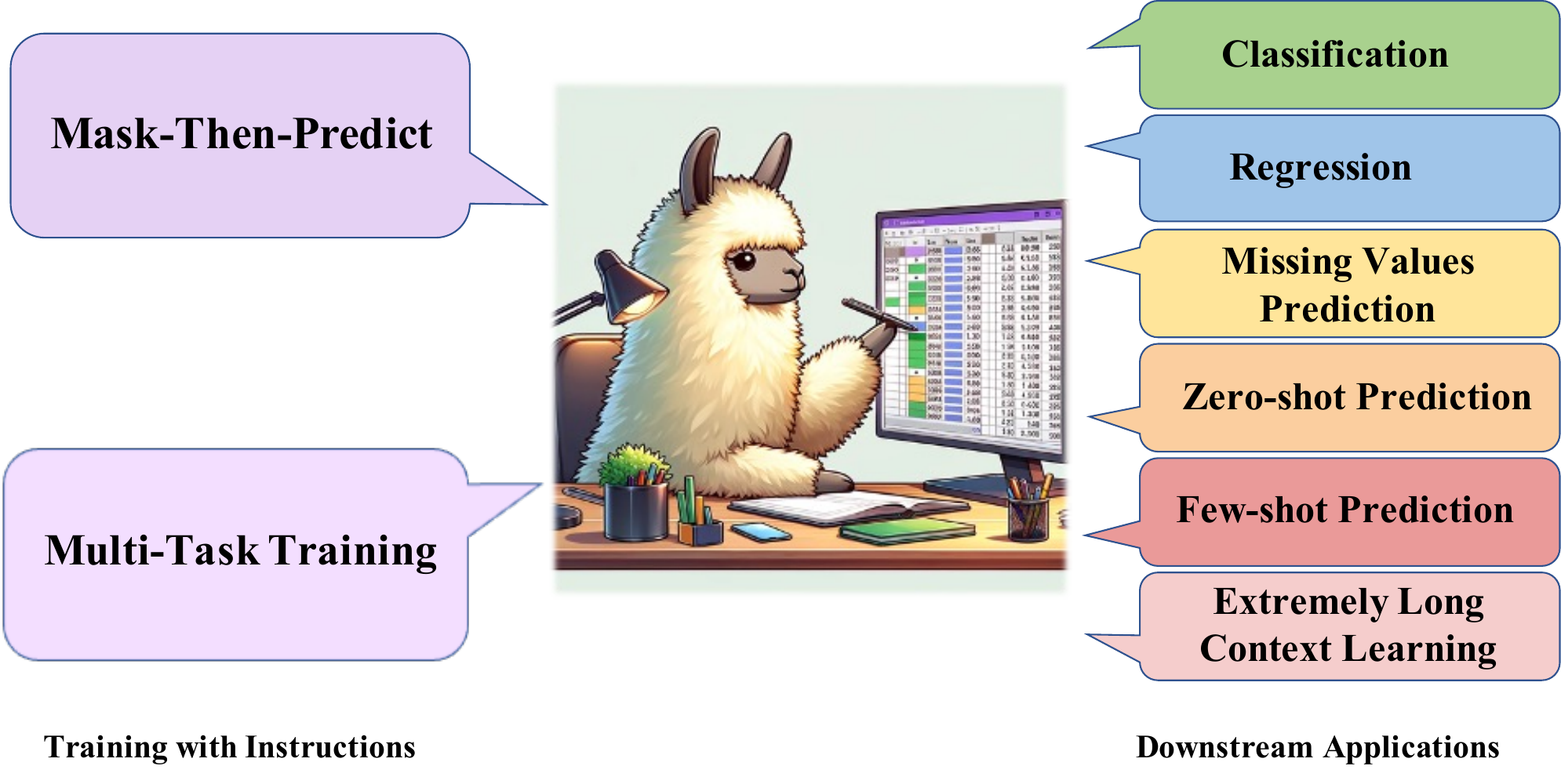}
  \caption{Illustration of our methodology of unlocking the potential of existing Large Language Models (LLMs) through training with large-scale tabular data and its subsequent application to downstream tasks. The left part demonstrates the paradigm of adapting existing LLMs through continued training on tables, while the right shows the common usage on downstream scenarios.}
  \label{fig/training_and_application}
\end{figure}

When addressing predictive tasks on tabular data, the modeling complexity arises from the multidimensional interactions and structural intricacies inherent to tables, which differ substantially from natural language and pose challenges for capturing their embedded semantics. Prior efforts to mitigate these challenges face several limitations: 1) Traditional approaches often depend on extensive feature engineering, including embedding layers that map continuous features into vector representations~\cite{gorishniy2022embeddings,wang2022transtab}and textualization techniques that convert tabular inputs into natural language~\cite{liu2022ptab,hegselmann2023tabllm,wang2024chain}. However, such task-specific feature engineering is labor-intensive and frequently relies on human-designed assumptions, introducing biases that hinder model generalization. 2) Although Large Language Models (LLMs) have achieved strong performance across a wide range of applications~\cite{gong2020tablegpt,li2023table,zhao2023investigating}, their ability to process structured tabular data remains limited due to the absence of specialized pretraining. Because LLMs are primarily trained on natural language corpora, they struggle to capture the intricate relational patterns characteristic of tabular structures. 3) While recent work has explored continued pretraining of LLMs on table-specific data~\cite{gong2020tablegpt,li2023table,zhang2023tablellama,zhao2023investigating}, these efforts predominantly emphasize text generation tasks such as table-based question answering (QA)~\cite{nan2022fetaqa,cheng2021hitab}, relation extraction~\cite{deng2022turl}, and cell-level content description~\cite{parikh2020totto,zhang2023tablellama}. Yet, these approaches largely overlook core data-science tasks, including classification, regression, and missing value imputation, that are central to practical tabular prediction problems. Moreover, the absence of a dedicated large-scale pretraining corpus tailored to data-science applications further restricts the applicability and generalizability of existing methods~\cite{herzig2020tapas,yin2020tabert,zhu2023xtab,slack2023tablet}.

To address the limitations of existing approaches, this work investigates the adaptation of large language models (LLMs) to predictive tabular tasks through large-scale, domain-aligned continued pretraining. We construct an extensive corpus of real-world tables, primarily sourced from Kaggle, spanning over 300 application domains and comprising approximately 13 billion examples. Unlike prior table-pretraining efforts that focus mainly on text-generation–oriented tasks such as table QA or semantic parsing, our dataset is tailored for predictive modeling in data science, characterized by numeric-heavy distributions, heterogeneous schemas, and diverse tabular structures. This broad coverage enables the model to acquire transferable semantic and relational representations that generalize effectively across varied downstream predictive workloads.

Instead of relying on task-specific feature engineering, we adopt structured instruction templates to prompt existing LLMs. This removes the need for domain-dependent preprocessing pipelines and reduces the biases associated with manually designed features. The use of instruction-driven prompting also preserves the native generative interface of LLMs, which allows the model to leverage its reasoning capabilities, including chain-of-thought, when solving downstream predictive tasks. In addition, we introduce a unified pretraining framework built upon the self-supervised learning paradigm inherent to LLMs. The framework integrates tabular content with task-oriented instructions and enables the model to jointly learn tabular semantics, relational patterns, and predictive behaviors within a single generative architecture. A unified prompt schema is used consistently across pretraining, finetuning, and inference, which ensures stable task transfer and avoids format discrepancies between training and application stages.

Our pretraining formulation incorporates multiple complementary objectives, such as column-name prediction, numerical cell reconstruction, and textual cell content generation, within the mask-then-predict and multi-task learning setup. These objectives encourage fine-grained reasoning over tabular structure and values while exploiting LLMs’ capacity to interpret natural language and human intent. By continuing pretraining on large-scale tabular data, the resulting models learn to extract informative features automatically and capture complex patterns, which substantially improves performance on classification, regression, and missing-value imputation. Figure~\ref{fig/training_and_application} provides an overview of the training methodology and its downstream applications.

Our exploration of large-scale tabular pretraining and its application to predictive tasks yields several key contributions:
\begin{itemize}
    \item We propose a simple yet effective pretraining strategy that repurposes LLMs from general-purpose text generators into predictive learners for structured tabular inputs. This framework differs from prior table-focused LLM work by directly addressing the central predictive tasks in data science, including classification, regression, and missing-value imputation, instead of generative QA or semantic parsing.
    \item We curate a large-scale dataset that provides structural and semantic diversity. This supports strong generalization and serves as a valuable resource for tabular modeling research with LLMs.
    \item Our trained model achieves superior performance, validated through extensive experimental evaluations across 30 classification and regression tasks. Compared to baseline methods, our approach yields an average improvement of 8.9\% in classification tasks and 10.7\% in regression tasks. For missing value prediction, our model surpasses GPT-4 by 27\%. Moreover, it exhibits a 28.8\% improvement in extreme-few-shot (4-shot) predictions across diverse datasets and an 18.8\% enhancement in long-context learning tasks. In addition, we access the generalizability of our approach through applying it to multiple open-sourced LLMs.
\end{itemize}

\begin{figure*}[ht]
\begin{center}
\centerline{\includegraphics[width=2.1\columnwidth]{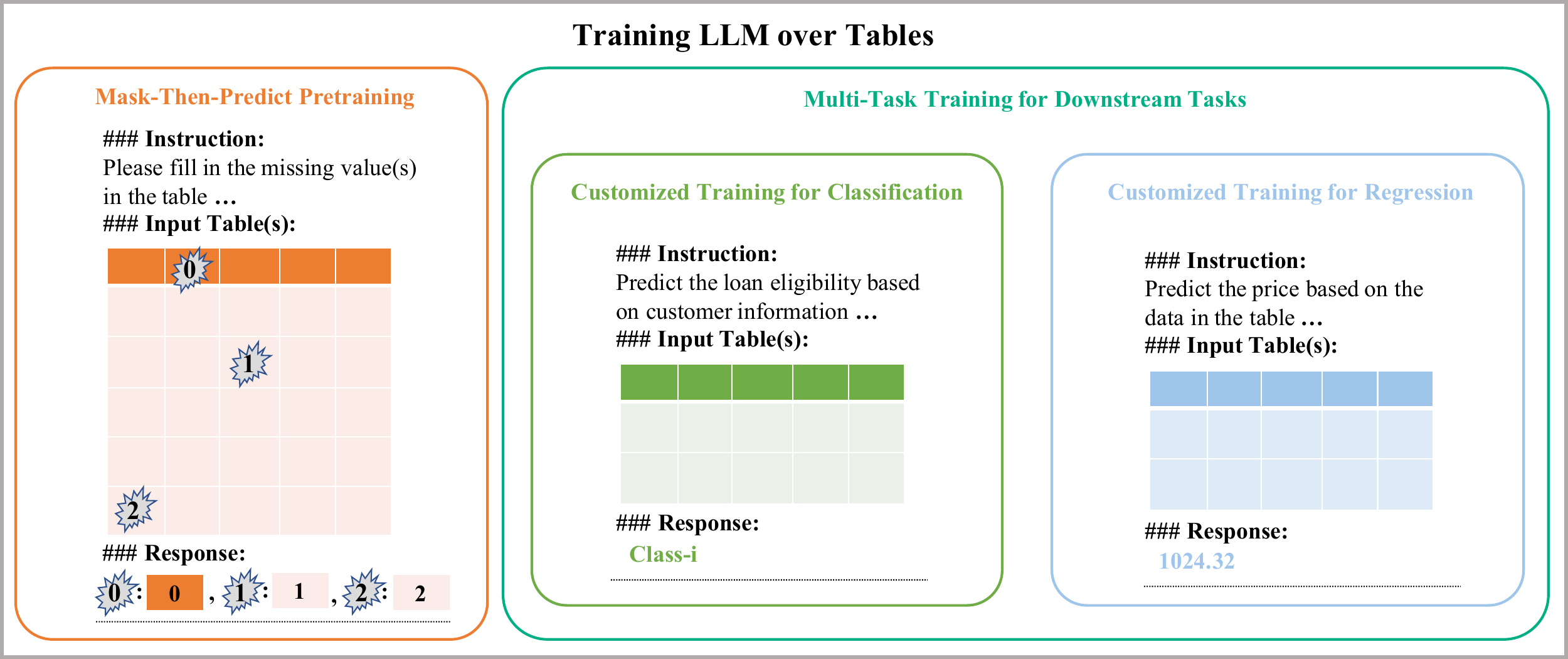}}
\caption{Illustration of the initial pretraining phase of a LLM applying the Mask-Then-Predict strategy (on the left), followed by the multi-task training phase customized for downstream tasks such as classification and regression (on the right). Through the former phase, the LLM acquires unstructured knowledge embedded within tables. Subsequently, during the latter phase, it enhances its capability for reasoning between instructions and tabular contents, curated for downstream tasks.}
\label{fig/overall_arc}
\end{center}
\end{figure*}

\section{Related Works}
\label{relw_sect}
\subsection{Traditional Machine Learning and Deep Learning Approaches for Tabular Data}
Tabular data modeling has traditionally been dominated by classical machine learning methods, particularly tree-based ensembles such as Random Forests and XGBoost~\cite{chen2016xgboost}. These models deliver strong predictive performance and remain competitive due to their interpretability and efficiency. However, their dependence on manual feature engineering and limited scalability in high-dimensional spaces has motivated the exploration of deep learning (DL) approaches. Recent DL methods aim to improve tabular modeling by integrating neural architectures. NODE~\cite{popov2019neural}, for example, combines decision-tree structures with dense neural connections, enabling gradient-based optimization while retaining hierarchical feature learning. Although effective, such hybrid models introduce additional complexity and computational overhead, reducing their adaptability across diverse tasks.

Transformer-based architectures~\cite{vaswani2017attention} have gained traction in tabular learning for their ability to capture long-range dependencies and relational patterns. Several works adapt Transformers for structured inputs. TabTransformer~\cite{huang2020tabtransformer} incorporates column embeddings to encode column-wise semantics, while FT-Transformer~\cite{gorishniy2021revisiting} employs a feature tokenizer to embed tabular features before Transformer processing. More recently, UniTabE~\cite{yang2024unitabe} proposes a unified encoder for heterogeneous table structures. Despite promising results, these models often require substantial feature engineering or domain-specific architectural design, which limits their generalizability across datasets and application settings.

\subsection{Large Language Models for Tabular Data}
The increasing capabilities of Large Language Models (LLMs) have motivated substantial interest in their application to tabular data tasks. TaBERT~\cite{yin20tabert} linearizes tables into text sequences and jointly encodes structured tables and natural language utterances through specialized feature engineering. TUTA~\cite{wang2021tuta} adopts a bi-dimensional hierarchical encoder that models cell coordinates and structural dependencies. XTab~\cite{zhu2023xtab} extends Transformer-based architectures with federated learning, combining dataset-specific and shared Transformer blocks to support collaborative tabular modeling.

A prominent line of work relies on textualizing table contents for LLM processing. TableFormer~\cite{yang2022tableformer} introduces a structurally aware encoder with learnable attention biases, making it invariant to row and column order and improving table reasoning performance. TabLLM~\cite{hegselmann2023tabllm} converts table rows into natural language using handcrafted or LLM-generated templates, enabling tabular tasks without additional finetuning. TabPFN~\cite{hollmann2022tabpfn} applies in-context learning~\cite{brown2020language} by formatting tabular inputs as text sequences for classification. Although these methods support zero-shot and few-shot inference, they remain fundamentally limited by the absence of domain-aligned pretraining on large-scale tabular corpora. Without such pretraining, LLMs struggle to generalize across heterogeneous tabular datasets and exhibit limited effectiveness in predictive modeling tasks that require fine-grained numerical and relational understanding.

\subsection{Pretraining for Tabular Data}
Several prior studies have explored pretraining on tabular data, but they primarily target text-generation tasks rather than predictive modeling. Methods such as PASTA~\cite{gu2022pasta}, TableLlama~\cite{zhang2023tablellama}, and Chain-of-Table~\cite{wang2024chain} focus on language-oriented objectives, including table-based question answering, semantic parsing, and table manipulation. For example, TaPas~\cite{herzig2020tapas} is designed for table QA, PASTA~\cite{gu2022pasta} improves Transformer-based table operations on cloze-style tasks, and TableLlama~\cite{zhang2023tablellama} adapts LLaMA-2 for table text generation without addressing predictive tasks.

In contrast, our approach is explicitly designed for predictive tabular modeling through large-scale LLM pretraining. We introduce three key advances: 1) we shift the focus from text-generation objectives to predictive tasks in data science; 2) we evaluate our method across multiple open-source LLMs, including Qwen-2.5, LLaMA-2, LLaMA-3.1, and DeepSeek-R1-distilled, demonstrating consistent generalization beyond a single architecture; and 3) we construct a large-scale corpus of 13 billion tabular examples spanning 300 domains, providing broad coverage for diverse real-world tasks. Our experiments show that continued pretraining enables LLMs to generalize across a wide range of tabular predictive tasks, outperforming existing baselines and illustrating the effectiveness of instruction-aligned pretraining for data science applications.

\section{Methodology}
\label{method_sect}
This section presents the key components of our methodology, including prompting LLMs with tabular data (\S\ref{subsec_unified_serialize}), continued pretraining on tables (\S\ref{sec_pretrain}), constructing the training corpus (\S\ref{sec_dataset_collection}), and applying the trained model to downstream tasks (\S\ref{sec_app_downstream_task}). Figure~\ref{fig/overall_arc} illustrates our training regimen, beginning with a Mask-Then-Predict pretraining task to capture unstructured knowledge from tables, followed by multi-task training tailored for downstream applications, including both classification and regression.

\begin{figure}[t!]
  \centering
  \includegraphics[width=0.9\columnwidth]{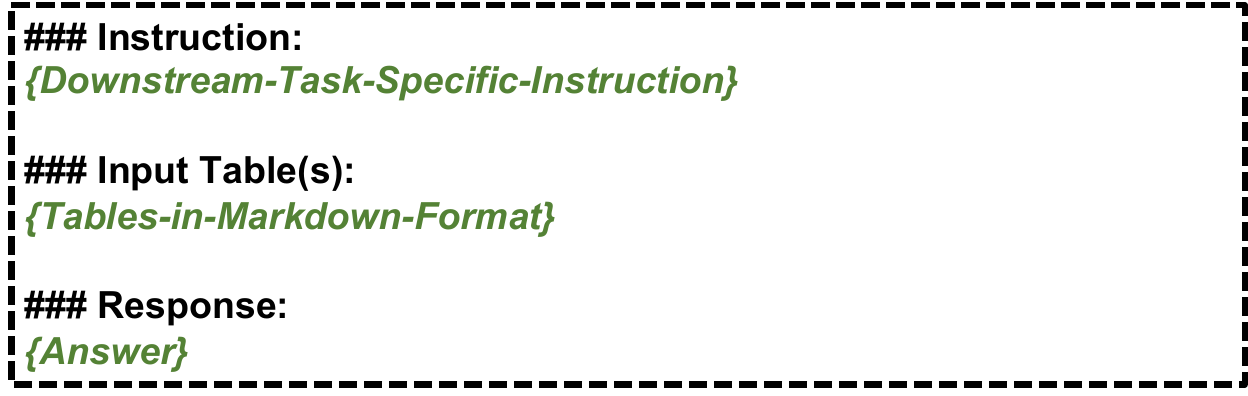}
  \caption{The unified prompt template used for combining the instruction with tables to form the model input in both pretraining and finetuning in downstream tasks.}\label{fig/unified_prompt_template}
\end{figure}

\subsection{Unified Serialization} 
\label{subsec_unified_serialize}
Inspired by recent findings~\cite{shin2023arxiveri}, which highlight the effectiveness of Markdown over conventional tabular formats such as CSV and HTML, we choose to serialize tables in Markdown. This decision is further motivated by Markdown’s ability to preserve table structure directly while accommodating diverse layouts with minimal feature engineering. 
In addition, we adopt a unified prompt template (Figure~\ref{fig/unified_prompt_template}) that integrates task-specific instructions with the table content. This structured design simplifies the model’s interpretation of the instructions and facilitates reasoning over the combined textual and tabular inputs. Moreover, the unified template enables consistent processing across tasks, thereby improving generalization and effectiveness in handling diverse tabular prediction settings.

\subsection{Two Stage Training with Tables}\label{sec_pretrain}
To pretrain the model, we employ a Mask-Then-Predict objective, analogous to the Masked Language Modeling (MLM) paradigm~\cite{devlin2018bert} widely used in NLP. This objective strengthens the model’s contextual understanding and its ability to capture relational dependencies within tabular data. We further incorporate multi-task training aligned with downstream applications, endowing the model with task-specific knowledge relevant to tabular prediction. By combining these objectives, our approach jointly promotes generalization, contextual reasoning, and task-aware capability, ultimately yielding a more versatile and effective model for diverse downstream tasks.

\begin{figure*}[t!]
\begin{center}
\centerline{\includegraphics[width=2.1\columnwidth]{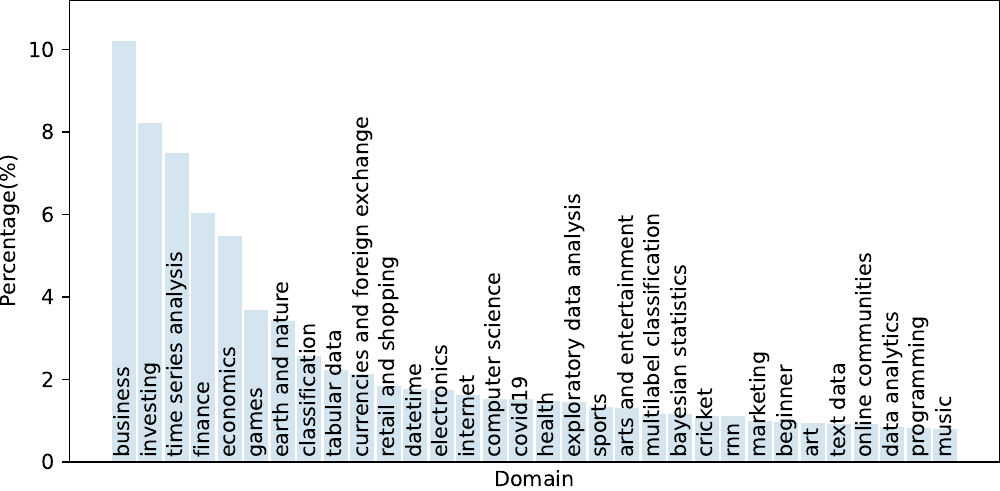}}
\caption{Top 32 domains in our Kaggle-collected pretraining corpus (around 300 domains, 13B tables). The dataset spans both numerical and textual columns, offering broad coverage of real-world tabular scenarios and enabling the model to learn diverse quantitative and natural language patterns.}
\label{fig/top32_domain_distribution}
\end{center}
\end{figure*}

\textbf{Mask-Then-Predict Pretraining} 
Our Mask-Then-Predict strategy randomly masks regions within input tables, prompting the model to infer the obscured values from surrounding context. This objective enables the model to effectively acquire structural and semantic knowledge from tabular data. Moreover, our method adopts a unified pretraining framework that spans diverse tasks, including predicting column names, numerical values, and textual cell contents, as illustrated on the left side of Figure~\ref{fig/overall_arc}. Engaging the model in such varied prediction tasks encourages a deeper understanding of table structure, semantics, and inter-cell relationships. Motivated by the observation that each table cell typically encodes a coherent piece of information, often representing a integrated entity or key attribute, we treat entire cells as the masking unit rather than individual tokens typically used in the pretraining of LLMs.

n practical applications, tables often contain multiple missing values. To model this scenario, we dynamically vary the number of masked cells during pretraining. The left portion of Figure~\ref{fig/overall_arc} presents an example with three masked cells. Each masked cell is replaced by a distinct sentinel token (for example, ``\textless missing\_value\_0\textgreater'', ``\textless missing\_value\_1\textgreater'', ..., ``\textless missing\_value\_\{N-1\}\textgreater''), allowing the model to distinguish among multiple masked positions within the same table.

\textbf{Multi-Task Training for Downstream Task} 
In addition to acquiring intra-table knowledge through Mask-Then-Predict pretraining, we further specialize the model using a multi-task training objective. This objective leverages datasets tailored for classification and regression, aiming to strengthen the model’s ability to reason over tabular content under task-specific instructions. Importantly, the training datasets are fully distinct from those used in our evaluation experiments. To enrich the training corpus, we annotate the regression and classification datasets with instructional prompts. Exposing the pretrained model to these diverse problem-solving scenarios, which closely resemble downstream applications, encourages a more comprehensive and adaptable understanding of task-specific requirements. 
Furthermore, our model is designed to predict full text sequences rather than single numerical values or class probability distributions. This design choice eliminates the need for task-specific prediction heads atop the LLM, allowing the model to address a wide range of prediction tasks directly and uniformly. This strategy highlights the flexibility and robustness of our approach in handling diverse predictive tasks in data science.

\begin{table}[tb]
\small 
\caption{\label{tbl_pretrain_ds_statistic} Comparison of dataset statistics between our training corpus and prior pretraining datasets, showing our dataset's larger sample size and broader domain coverage.
  }
  \centering
  \scalebox{1.0} {
  \begin{tabular}{l c c}
    \toprule
    \textbf{Method} & \textbf{\#Examples} & \textbf{Sources} \\ 
    \midrule 
    TUTA\cite{wang2021tuta} & 58M & WikiTable,WDC,web \\
    TAPAS\cite{herzig2020tapas} & 6.2M & WikiTable \\
    TaBERT\cite{yin2020tabert} & 26.6M & WikiTable, WDC \\
    XTab\cite{zhu2023xtab} & 52Tasks & OpenML-AutoML \\
    Ours & 13B & UCI,Kaggle (300 domains) \\
  \bottomrule
  \end{tabular}
  }
\end{table}

\subsection{Pretraining Data Collection}\label{sec_dataset_collection}
Our pretraining dataset, sourced from Kaggle, spans 300 domains and contains 13 billion examples. This breadth of domain coverage positions our model for strong domain independence and enhances its versatility. In this work, we focus exclusively on tabular data, categorizing columns into numerical and textual types commonly encountered in data science, while excluding image, audio, and video modalities. Numerical columns constitute the majority (approximately 60.5\%), reflecting the prevalence of quantitative information such as integers, decimals, and percentages, whereas textual columns account for nearly 40\%. This balanced composition enables the model to develop a well-rounded understanding of both quantitative structure and natural language semantics, rather than being dominated by either type alone. Models trained on such diverse data demonstrate improved versatility and effectiveness in addressing real-world problems that demand comprehensive data interpretation and analysis.

We further augment the corpus with multi-task training data designed to adapt the LLM to downstream tabular tasks. This includes classification and regression datasets from the UCI Machine Learning Repository\footnote{https://archive.ics.uci.edu/datasets}, each annotated with task-specific instructions. A complete list of datasets is provided in the Appendix. Table~\ref{tbl_pretrain_ds_statistic} reports comparative statistics between our dataset and those used in previous work, highlighting its broad domain coverage and large-scale composition.

Regarding licensing, all Kaggle data collection strictly adheres to the platform’s terms of use and dataset-level licensing requirements. We exclusively selected datasets accessible through the official Kaggle API\footnote{https://github.com/Kaggle/kaggle-api} that are publicly available and compatible with open-source redistribution. These include datasets licensed under the GNU General Public License (GPL) and the Open Data Commons suite (ODbL, PDDL, ODC-By). This careful selection ensures that our model can be openly shared. For example, datasets such as ``sales-from-different-stores'', obtained via the official API and distributed under the ODbL license, permit worldwide and royalty-free use.

\subsection{Applications in Downstream Tasks}\label{sec_app_downstream_task}
This section explores the application of our trained model to various downstream tasks, such as filling in missing values, performing classification and regression, and executing broader tasks like zero-shot and in context learning.

\noindent\textbf{Finetuning for Classification.} 
Similar to traditional finetuing LLM for classification tasks, an additional classification head is integrated into our trained model. The model is optimized to minimize the cross-entropy loss, ensuring a more accurate alignment with the ground truth data. For the few-shot prediction, our trained model is finetuned with limited training examples or data points.

\noindent\textbf{Finetuning for Regression.} 
In a manner akin to classification, the model is augmented with an additional regression head for this task. The focus of optimization shifts to reducing the mean squared error (MSE) between the model's predictions and the actual values. 
Apart from adding a head layer, the model can be finetuned for classification or regression by predicting target values through text generation.

\noindent\textbf{Finetuning for Missing Values Prediction.} 
Handling missing values is a common yet critical challenge in real-world tabular applications. By applying the same methodology used in our mask-then-predict pretraining, the model is capable of predicting missing values at each designated sentinel position. This approach leverages the model’s learned representations from large-scale table pretraining to infer and generate contextually appropriate content, thereby improving the integrity and completeness of the imputed tables.

\noindent\textbf{In Context Learning} 
Before predicting a target example, the model is provided with several contextual examples. These examples are formatted using the unified prompt template, ensuring consistency in the input structure. The texts from both the contextual examples and the target example are then concatenated into a single input sequence, enabling the model to infer the target prediction based on the given context.

\noindent\textbf{Zero-shot Prediction} 
The model can perform predictions on new examples without any finetuning. For classification, constrained decoding can be applied to select outputs from predefined token options.

\section{Experiments}
\label{exp_ana_sect}
This section presents a comprehensive evaluation of our model’s performance across a range of tasks. We begin with an overview of the experimental setup in Section~\S\ref{subsec_exp_setup}, followed by the results on downstream tasks in Section~\S\ref{subsec_overall_results}. Section~\S\ref{subsec_analysis} provides an in-depth analysis highlighting the model’s strengths and limitations. We further examine the application of our pretraining framework to a broader set of open-source LLMs in Section~\S\ref{subsec_pretrain_more_llms}. Finally, Section~\S\ref{subsec_case_study} presents a case study demonstrating the practical utility of our approach.

\subsection{Experimental Setup}
\label{subsec_exp_setup}
This section describes the experimental framework used to evaluate our pretrained LLM on tabular predictive tasks. We begin by detailing the implementation, including model architecture, training configurations, and computational resources. Next, we introduce the comparison baselines, which encompass traditional machine learning models, existing LLM-based approaches, and state-of-the-art methods. We then present the benchmark datasets used for evaluation, selected to provide a diverse and comprehensive assessment. Finally, we define the key metrics employed to measure model performance across classification, regression, and missing value prediction tasks.

\textbf{Implementation Details} 
We employ the Llama-2 7B as the foundational architecture for our model, using a high-performance computing cluster with NVIDIA A100 GPUs for efficient and capable training. We initiate training with a learning rate of \(2e{-5}\), balancing convergence speed with model stability. To accommodate large batch sizes, we use gradient accumulation with a step size of 4. The Adam optimizer, with hyperparameters \(\beta_1=0.9\), \(\beta_2=0.95\), and \(\epsilon=10^{-8}\), is adopted to ensure smooth and stable training progression. Following the masking recipe of BERT, a masking ratio of 0.15 is applied, randomly selecting cells for masking to bolster the model's competence with incomplete data. A warm-up ratio of 0.05 during the initial training phase helps prevent early instability by gradually adjusting the learning rate. 
In addition, numerical values are standardized to a precision of five decimal places to prevent excessively long numerical tokens. Tabular data can contain a wide range of information across multiple rows and columns, leading to the long sequence of model input. The context length determines how much of this data can be considered in a single model prediction. Inspired by recent research~\cite{xiong2023effective, fu2024data}, we have adjusted the base of RoPE (Rotary Positional Embedding) to enhance the model's ability to manage longer contextual dependencies. Furthermore, we further examine our proposed method with other open source models, including Qwen-2.5 7B~\footnote{https://huggingface.co/Qwen/Qwen2.5-7B-Instruct}, Llama-3.1 8B~\footnote{https://huggingface.co/meta-llama/Llama-3.1-8B}, and DeepSeek-R1-distilled Llama-3.1 8B~\footnote{https://huggingface.co/deepseek-ai/DeepSeek-R1-Distill-Llama-8B}.

\textbf{Baselines} 
In this work, we adopt the XGBoost as the representative baseline for traditional tree-based methods. The configuration for XGBoost adheres to the default settings of the xgboost package.\footnote{https://xgboost.readthedocs.io/en/stable/python/python\_api.html} For preprocessing textual data for XGBoost, we employed one-hot encoding through the scikit-learn library.\footnote{http://scikit-learn.org/stable} We also compare with GANDALF~\cite{joseph2023gandalf} that builds upon a tailored tabular processing unit combined with a gating mechanism and in-built feature selection. 
Furthermore, our comparative analysis incorporates a comprehensive range of Transformer-based models and pre-trained models, including Tapas~\cite{herzig2020tapas}, TaBERT~\cite{yin20tabert}, TabTransformer~\cite{huang2020tabtransformer}, TabPFN~\cite{hollmann2022tabpfn}, TUTA~\cite{wang2021tuta}, TabLLM~\cite{hegselmann2023tabllm}, and XTab~\cite{collins2022fedavg}, among others. Throughout the training of these models on downstream tasks, we follow the official hyperparameter settings to ensure a consistent and fair comparison. 
Note that models initially not designed to support regression tasks in their official code, such as TUTA, are adapted by just modifying the output layer to produce a single numerical output.

\noindent\textbf{Benchmarks} 
We curated a collection of datasets to thoroughly evaluate our proposed method against existing approaches. The collection includes four classification and three regression tasks, all sourced from Kaggle. Additionally, we incorporate tasks from publicly available tabular benchmarks~\cite{grinsztajn2022tree}\footnote{https://huggingface.co/datasets/inria-soda/tabular-benchmark}. Within this subset of benchmarks, we exclude tasks considered too easy to ensure that our evaluation accurately reflects the model's capabilities across a range of challenges. 
For tasks within the public tabular benchmarks, an abbreviated task name prefixed with ``n'' indicates that the dataset contains only numerical features, whereas a prefix of ``c'' denotes datasets with both textual and numerical features, providing a more complex evaluation scenario. This differentiation enables a detailed analysis of model performance across diverse data compositions commonly encountered in real-world applications. Importantly, there is no overlap between the pretraining data and the evaluation data for target tasks, preventing potential data leakage and ensuring a fair comparison.

\textbf{Metrics} 
The primary objective of this work is to further pretrain large language models on tabular data and evaluate their effectiveness on three core data science tasks: classification, regression, and missing-value imputation. To measure discriminative performance in classification, we report ROC-AUC. For regression, we adopt the coefficient of determination (\(R^2\)) as the evaluation metric. For missing-value prediction, which encompasses both textual and numerical entries, we evaluate textual values using ROUGE-L~\cite{lin2004rouge} and numerical values using \(R^2\).

\begin{table*}[htp]
  \caption{\label{tbl_cls_results_pub_ds} Evaluation results with ROC-AUC on classification tasks from Kaggle (left section) and public tabular benchmarks (right section). A higher score reflects superior results. The Best resutls in the table are denoted by bold formatting. The task name of each public benchmark starting with ``n'' represents the dataset only contains numerical features, while the task name starting with ``c'' denotes its dataset has both textual and numerical features. 
  Left section demonstrates the results of tabular tasks from Kaggle. 
  }
  \centering
  \resizebox{\linewidth}{!}{
  \begin{tabular}{l c c c c| c c c c c c c c c c c c c}
    \toprule
    \textbf{Method/Dataset} & \textbf{loan} & \textbf{heart} & \textbf{health} & \textbf{diabetes} & \textbf{cAlbe} & \textbf{cCTY} & \textbf{cDCCC} & \textbf{cElec} & \textbf{cRS} & \textbf{nDiab} & \textbf{nMT} & \textbf{nBM} & \textbf{nCT} & \textbf{nCred} & \textbf{nDCCC} & \textbf{nElec} & \textbf{nHelo} \\ 
     \midrule 

    XGBoost & 0.733 & 0.829 & 0.854 & 0.793 & 0.700 & 0.674 & 0.753 & 0.963 & 0.879 & 0.631 & 0.936 & 0.846 & 0.935 & 0.840 & 0.751 & 0.952 & 0.764 \\
    NODE & 0.712 & 0.789 & 0.845 & 0.821 & 0.705 & 0.711 & 0.758 & 0.868 & 0.812 & 0.634 & 0.925 & 0.845 & 0.827 & 0.807 & 0.764 & 0.855 & 0.766 \\
    AutoInt & 0.663 & 0.801 & 0.846 & 0.814 & 0.689 & 0.710 & 0.463 & 0.585 & 0.501 & 0.499 & 0.895 & 0.854 & 0.502 & 0.758 & 0.773 & 0.838 & 0.500 \\
    Tapas & 0.710 & 0.829 & 0.825 & 0.788 & 0.685 & 0.702 & 0.723 & 0.973 & 0.867 & 0.618 & 0.931 & 0.853 & 0.938 & 0.811 & 0.724 & 0.959 & 0.744 \\
    TaBERT & 0.666 & 0.741 & 0.819 & 0.788 & 0.704 & 0.692 & 0.763 & 0.965 & 0.519 & 0.627 & 0.928 & 0.857 & 0.955 & 0.823 & 0.730 & 0.952 & 0.763 \\
    TabTransformer & 0.580 & 0.811 & 0.838 & 0.806 & 0.441 & 0.697 & 0.722 & 0.821 & 0.733 & 0.623 & 0.852 & 0.821 & 0.654 & 0.740 & 0.431 & 0.819 & 0.505 \\
    FT-Transformer & 0.488 & 0.794 & 0.831 & 0.805 & 0.654 & 0.535 & 0.497 & 0.887 & 0.844 & 0.640 & 0.932 & 0.836 & 0.913 & 0.815 & 0.778 & 0.879 & 0.538 \\
    TabNet & 0.711 & 0.684 & 0.841 & 0.781 & 0.501 & 0.607 & 0.419 & 0.830 & 0.497 & 0.533 & 0.547 & 0.759 & 0.903 & 0.815 & 0.480 & 0.852 & 0.770 \\
    TUTA & 0.728 & 0.695 & 0.836 & 0.824 & 0.696 & 0.614 & 0.748 & 0.487 & 0.571 & 0.633 & 0.898 & 0.814 & 0.737 & 0.734 & 0.756 & 0.518 & 0.617 \\
    TabPFN & 0.710 & 0.787 & 0.800 & 0.821 & 0.703 & 0.697 & 0.762 & 0.859 & 0.782 & 0.632 & 0.923 & 0.849 & 0.846 & 0.838 & 0.767 & 0.858 & 0.721 \\
    XTab & 0.722 & 0.824 & 0.854 & 0.827 & 0.708 & 0.704 & 0.761 & 0.902 & 0.881 & 0.641 & 0.928 & 0.858 & 0.954 & 0.825 & 0.762 & 0.886 & 0.784 \\
    GANDALF & 0.646 & 0.796 & 0.822 & 0.819 & 0.704 & 0.699 & 0.693 & 0.820 & 0.822 & 0.635 & 0.924 & 0.847 & 0.828 & 0.792 & 0.496 & 0.847 & 0.775 \\
    TabLLM & 0.732 & 0.783 & 0.836 & 0.790 & 0.650 & 0.691 & 0.719 & 0.861 & 0.849 & 0.622 & 0.799 & 0.839 & 0.790 & 0.788 & 0.713 & 0.858 & 0.762 \\
    Llama2 7B & 0.706 & 0.774 & 0.841 & 0.817 & 0.687 & 0.683 & 0.711 & 0.962 & 0.883 & 0.573 & 0.893 & 0.815 & 0.954 & 0.802 & 0.736 & 0.964 & 0.764 \\
     \midrule 
    Our Method & \textbf{0.780} & \textbf{0.841} & \textbf{0.868} & \textbf{0.854} & \textbf{0.724} & \textbf{0.715} & \textbf{0.781} & \textbf{0.986} & \textbf{0.921} & \textbf{0.655} & \textbf{0.954} & \textbf{0.873} & \textbf{0.982} & \textbf{0.851} & \textbf{0.791} & \textbf{0.985} & \textbf{0.793} \\
  \bottomrule
  \end{tabular}}
\end{table*}

\begin{table*}[t]
  \caption{\label{tbl_reg_results_ds} Regression performance ($R^2$) on Kaggle datasets (left) and public tabular benchmarks (right). The task name of each public benchmark starting with `n'' represents the dataset only contains numerical features, while the task name starting with c'' denotes its dataset has both textual and numerical features. Higher values indicate better performance. Our method consistently outperforms prior approaches across diverse numerical and mixed-feature datasets.
  }
  \centering
  \scalebox{1.0} {
  \begin{tabular}{l c c c| c c c c c c c c c c}
    \toprule
    \textbf{Method/Dataset} & \textbf{LC} & \textbf{HP} & \textbf{PMI} & \textbf{cAbal} & \textbf{cAS} & \textbf{cHS} & \textbf{cNTGD} & \textbf{cPM} & \textbf{cSeat} & \textbf{nAbal} & \textbf{nElev} & \textbf{nH1} & \textbf{nHS} \\ 
     \midrule 

    XGBoost & 0.981 & 0.868 & 0.823 & 0.535 & 0.964 & 0.896 & 0.601 & 0.716 & 0.174 & 0.492 & 0.873 & 0.508 & 0.887 \\
    NODE & 0.967 & 0.883 & 0.856 & 0.523 & 0.938 & 0.803 & 0.464 & 0.641 & 0.074 & 0.491 & 0.862 & 0.404 & 0.802 \\
    AutoInt & 0.956 & 0.851 & 0.847 & 0.534 & 0.926 & 0.859 & 0.406 & 0.640 & 0.137 & 0.513 & 0.796 & 0.416 & 0.849 \\
    TaBERT & 0.880 & 0.808 & 0.784 & 0.418 & 0.915 & 0.607 & 0.665 & 0.667 & 0.122 & 0.447 & 0.839 & 0.419 & 0.527 \\
    TabTransformer & 0.974 & 0.847 & 0.668 & 0.517 & 0.427 & 0.745 & 0.328 & 0.496 & 0.126 & 0.504 & 0.691 & 0.185 & 0.717 \\
    FT-Transformer & 0.981 & 0.590 & 0.691 & 0.513 & 0.928 & 0.874 & 0.404 & 0.669 & 0.107 & 0.516 & 0.447 & 0.448 & 0.867 \\
    TabNet & 0.967 & 0.763 & 0.527 & 0.504 & 0.964 & 0.830 & 0.403 & 0.618 & 0.161 & 0.505 & 0.360 & 0.304 & 0.709 \\
    TUTA & 0.956 & 0.805 & 0.854 & 0.304 & 0.871 & 0.619 & 0.620 & 0.569 & 0.173 & 0.244 & 0.625 & 0.299 & 0.606 \\
    GANDALF & 0.962 & 0.864 & 0.845 & 0.521 & 0.944 & 0.878 & 0.331 & 0.636 & 0.157 & 0.513 & 0.856 & 0.291 & 0.869 \\
    Llama2 7B & 0.967 & 0.854 & 0.816 & 0.363 & 0.965 & 0.846 & 0.658 & 0.708 & 0.162 & 0.460 & 0.865 & 0.458 & 0.860 \\
    \midrule 
    Our Method & \textbf{0.985} & \textbf{0.890} & \textbf{0.874} & \textbf{0.552} & \textbf{0.981} & \textbf{0.901} & \textbf{0.745} & \textbf{0.721} & \textbf{0.182} & \textbf{0.532} & \textbf{0.895} & \textbf{0.530} & \textbf{0.892} \\
  \bottomrule
  \end{tabular}
  }
\end{table*}

\subsection{Downstream Task Evaluation} 
\label{subsec_overall_results}
\textbf{Classification } 
Table~\ref{tbl_cls_results_pub_ds} presents the comparative performance of various methods on classification tasks. The results demonstrate that our method outperforms traditional approaches (e.g., XGBoost), Transformer-based models, as well as pretrained models designed for tabular data (e.g., TaBERT, TUTA, XTab, TabLLM). Our approach consistently achieves superior performance across a wide range of classification tasks, with an average improvement of 8.9\%, highlighting its effectiveness for real-world applications. 
Furthermore, Table~\ref{tbl_cls_results_pub_ds} reveals a general trend: methods tend to exhibit lower performance on mixed-feature tasks compared to purely numerical tasks. This disparity is particularly pronounced in models such as TUTA. In contrast, our approach maintains high performance across all task types, illustrating the robustness of our pretraining strategy in handling both numerical and textual data, supported by a training corpus rich in both feature types.

\noindent\textbf{Regression } 
In the realm of regression, Table~\ref{tbl_reg_results_ds} reports the \(R^2\) metrics across multiple tasks. Consistent with its performance on classification tasks, our method outperforms previous approaches on both Kaggle regression datasets and other public tabular benchmarks. This demonstrates that our method generalizes effectively across a wide range of tasks, rather than being limited to in-domain scenarios. Overall, our model achieves an average performance improvement of 10.7\% on these regression tasks. The consistently high scores highlight our method's ability to capture complex patterns and relationships within tables, resulting in accurate regression predictions.

\begin{figure}[htb]
\begin{center}
\centerline{\includegraphics[width=0.96\columnwidth]{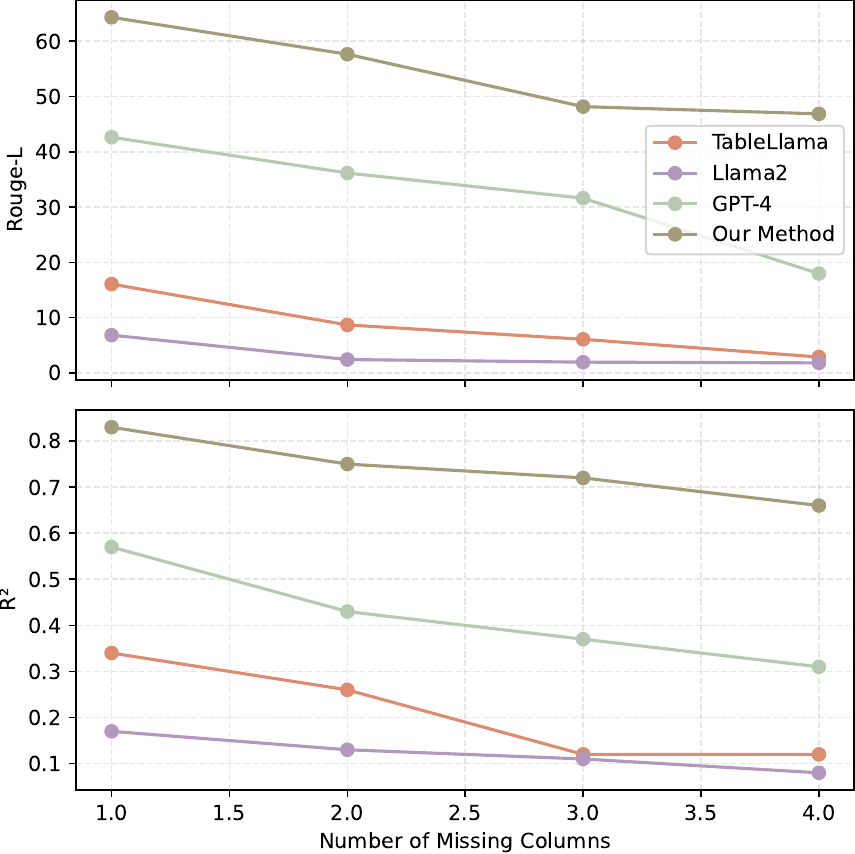}}
\caption{Comparison of missing value prediction with 1 to 4 concealed cells. Top: textual predictions evaluated with Rouge-L; bottom: numerical predictions evaluated with $R^2$. Our method consistently outperforms TableLlama, Llama-2, and GPT-4, with larger gains as the number of missing values increases, highlighting its effectiveness for data completion and recovery in tabular tasks.}
\label{fig/fill_in_missing_val_analysis}
\end{center}
\end{figure}

\noindent\textbf{Filling in Missing Values } 
Additionally, we evaluate the effectiveness of our method in filling in missing values. By simulating missing data through the random removal of cell content in tables, we tasked the model with predicting these absent values under varying conditions of data sparsity. The results are shown in Figure~\ref{fig/fill_in_missing_val_analysis}. The numerical values are measured with \(R^2\) metric, while the textual contents are evaluated with Rouge-L. Our model's performance is benchmarked against TableLlama~\cite{zhang2023tablellama}, Llama-2, and GPT-4. Despite TableLlama being an extension of Llama-2 and trained specifically on several tabular tasks (e.g. table-to-text, TableQA), its incremental performance gain over Llama-2 in handling missing values is modest, even in the prediction of textual values, indicating the necessity of training existing LLM with the mechanism of mask-then-predict while facing with missing contents. In contrast, our model demonstrates a clear improvement, particularly noteworthy as the number of missing values increases. Such improvement affirms our method's potential applications related to data completion, data recovery and tabular data synthesis.

\begin{figure*}[ht]
\begin{center}
\centerline{\includegraphics[width=2.1\columnwidth]{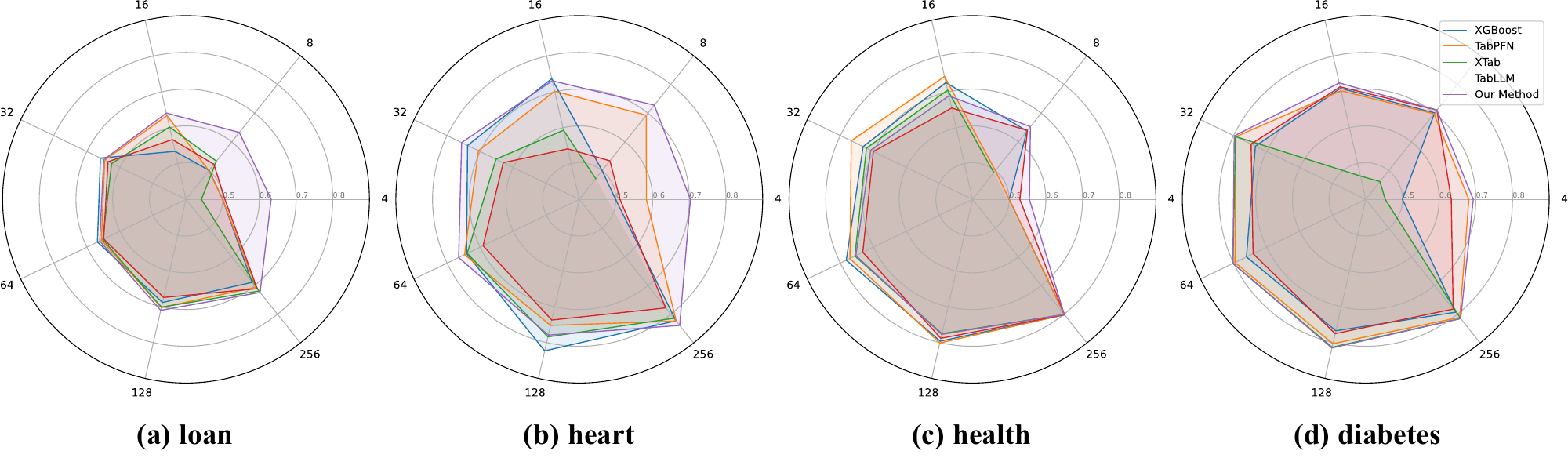}}
\caption{Radar chart illustrating the performance of few-shot prediction in 4 classification tasks. The evaluation metric is ROC-AUC. Our method demonstrates superior performance, achieving higher scores in most of the directions (number of shots) on the chart, showing its effectiveness and competitiveness.}
\label{fig/few_shot_comparison}
\end{center}
\end{figure*}

\begin{figure}[htb]
\begin{center}
\centerline{\includegraphics[width=1.0\columnwidth]{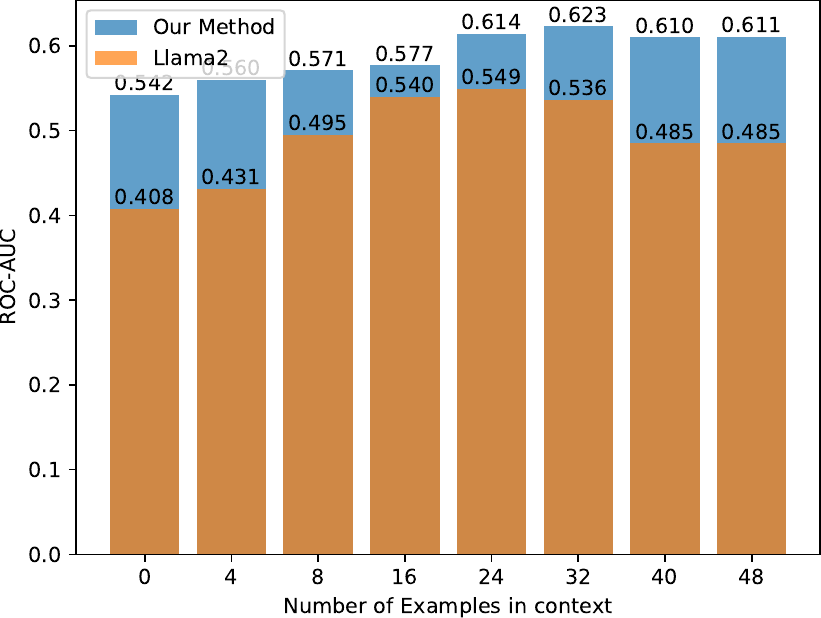}}
\caption{Analysis of extremely long context learning. The x-axis shows the number of examples in the context (from 0 to 48). The blue regions indicate where our method outperforms Llama-2, with larger blue areas reflecting greater improvements. }
\label{fig/long_context_analysis}
\end{center}
\end{figure}

\noindent\textbf{Extremely Long Context Learning } 
In tabular tasks, modeling long sequences presents a significant challenge, requiring the model to handle the wide-ranging complexity of the data while addressing issues such as data sparsity and non-linear relationships within tables. We further evaluate our model under an extreme long-context learning scenario. For each test sample, we select \(k\)-nearest examples from the training set as contextual input according to cosine similarity. Each example is converted into a vector representation using sentence-transformers\footnote{https://github.com/UKPLab/sentence-transformers} based on its natural language text. The content of each row is processed in the following format: ``column-name-0 is cell-value-0, column-name-1 is cell-value-1, ..., column-name-\{N-1\} is cell-value-\{N-1\}''. This conversion aids the LLM—especially sentence-transformers not specifically trained on tables—in distinguishing subtle differences among examples more effectively than using the original table format directly. To maintain label balance, an equal number of examples from each class are selected.

Figure~\ref{fig/long_context_analysis} shows the comparative performance of our model against Llama-2 7B 80K~\cite{fu2024data}, which supports up to 80K tokens. The results demonstrate a clear performance improvement, with an average gain of 18.8\%, indicating that our model not only achieves higher scores but also consistently surpasses Llama-2 7B 80K as the context size increases. This highlights our model's capability to handle extremely long-context learning effectively.

\noindent\textbf{Zero-shot Prediction} 
The results of zero-shot prediction are also presented in Figure~\ref{fig/long_context_analysis}. Our method achieves a ROC-AUC of 0.54 and an accuracy of 68\%, representing a substantial improvement over the baseline model. This performance gain highlights the effectiveness of our approach and demonstrates the model's ability to generalize from pretraining, applying its learned knowledge to new, unseen tasks without additional fine-tuning. The results indicate robust adaptability and the potential of our method to handle diverse predictive tasks with minimal prior information. Moreover, the consistent improvement observed as the number of context examples increases further illustrates the model's capacity to leverage additional contextual information for enhanced learning.

\begin{table}[t]
\small 
  \caption{\label{tbl_ablation_analysis} Ablation on classification (nDCCC) and regression (cHS) tasks. Removing Mask-Then-Predict greatly reduces performance, while omitting multi-task training has a smaller effect; removing both leads to the largest drop.
  }
  \centering
  \scalebox{1.0} {
  \begin{tabular}{l c c}
    \toprule
    \textbf{Method/Task} & \textbf{Classification} & \textbf{Regression} \\ 
    \midrule 
    Our Method & 0.791 & 0.901 \\
     \midrule 
    - w/o Mask-then-Predict & 0.754 & 0.865 \\
    - w/o Customized Tuning & 0.773 & 0.888 \\
    - w/o both objectives & 0.736 & 0.846 \\
  \bottomrule
  \end{tabular}
  }
\end{table}

\noindent\textbf{Few-shot Prediction } 
To evaluate our model in data-limited scenarios, we conducted few-shot prediction experiments. As shown in Figure~\ref{fig/few_shot_comparison}, our method achieves an average performance improvement of 28.8\% in extreme few-shot (4-shot) predictions across various datasets compared to baseline methods, highlighting its adaptability to new target domains with minimal data. This indicates that our pretrained model is highly data-efficient. 

The results further show that increasing the number of shots generally improves performance across all methods and datasets, although the extent of improvement varies. Performance differences between methods tend to diminish once the number of training examples exceeds 64. Nevertheless, our method continues to outperform competitors even with larger training sets (e.g., 128 and 256 examples). Notably, pretrained methods outperform XGBoost in the 4-shot setting, emphasizing the benefits of pretraining in leveraging tabular knowledge. Overall, these findings demonstrate that our approach is particularly effective in few-shot learning scenarios, deriving substantial gains from pretraining and providing a robust solution for predictive tasks with limited data.

\subsection{Analysis}
\label{subsec_analysis}
In this subsection, we conduct a detailed analysis of our model’s performance. First, we perform an \textit{Ablation Study} to quantify the contribution of key components. Next, we examine the impact of different \textit{Learning Rates} and evaluate the benefits of \textit{Finetuning with LoRA}. We also assess the model’s capability in \textit{Predicting Missing Values} and the effectiveness of \textit{Chain-of-Thought (CoT) Prompting}. Finally, we analyze robustness to \textit{Label Imbalance}, providing a comprehensive understanding of the model’s strengths and limitations across diverse conditions.

\textbf{Ablation Study} 
We examine the contributions of the key pretraining objectives in our model, as summarized in Table~\ref{tbl_ablation_analysis}. Removing the Mask-Then-Predict objective results in a substantial decrease in both classification and regression performance, highlighting its critical role in learning effective representations from tabular data. In contrast, omitting the multi-task training objective leads to a more modest performance decline, indicating that while it contributes to the model's effectiveness, its impact is less pronounced than that of the Mask-Then-Predict objective. Removing both objectives causes a significant drop in performance across all tasks, emphasizing the synergistic benefit of combining these objectives within our pretraining framework. These results underscore the importance of each objective in optimizing the model’s ability to handle structured tabular data effectively.

\begin{figure}[ht]
\begin{center}
\centerline{\includegraphics[width=0.95\columnwidth]{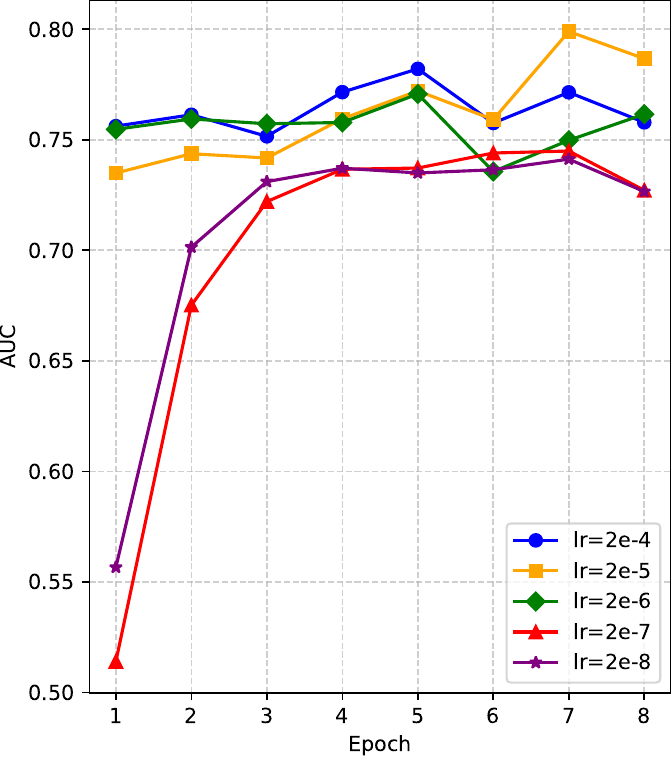}}
\caption{Impact of learning rate on model performance (AUC) across training epochs for the credit prediction task. Smaller learning rates show rapid initial gains, especially in the first four epochs. With an appropriately chosen learning rate, our model quickly acquires knowledge, and its performance reaches a high level in a short number of epochs.}
\label{fig/lr_analysis}
\end{center}
\end{figure}

\begin{figure}[ht]
\begin{center}
\centerline{\includegraphics[width=0.95\columnwidth]{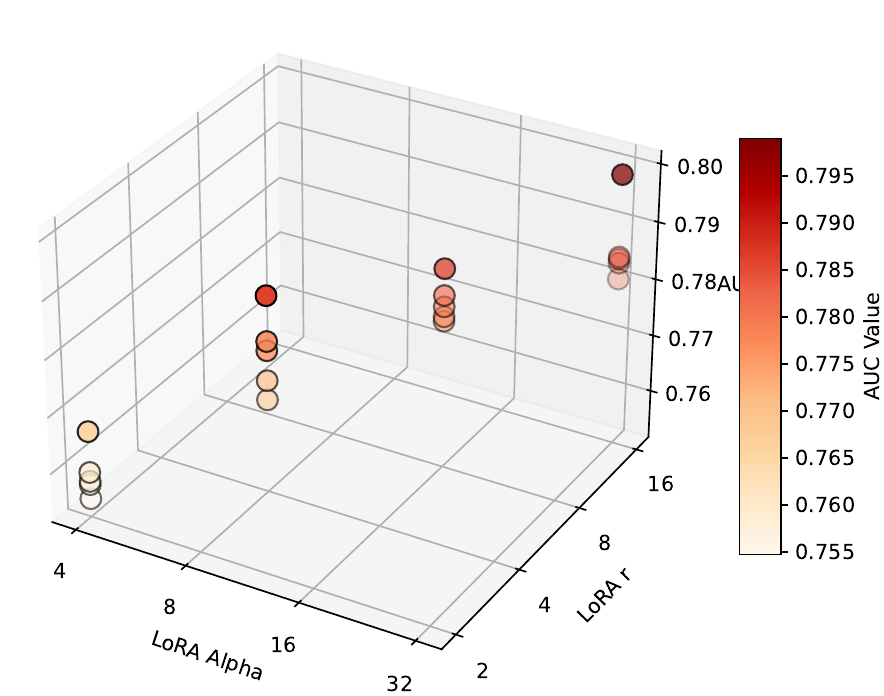}}
\caption{3D scatter plot of AUC values for different LoRA hyperparameter settings. The x-axis represents LoRA Alpha ($\alpha$), the y-axis represents LoRA Rank ($r$), and the z-axis shows the AUC performance. Each point corresponds to a specific ($\alpha$, $r$) combination, with color intensity indicating AUC magnitude (darker shades represent higher values). Each combination takes five runs.}
\label{fig/lora_analysis}
\end{center}
\end{figure}

\begin{figure}[ht]
\begin{center}
\centerline{\includegraphics[width=0.95\columnwidth]{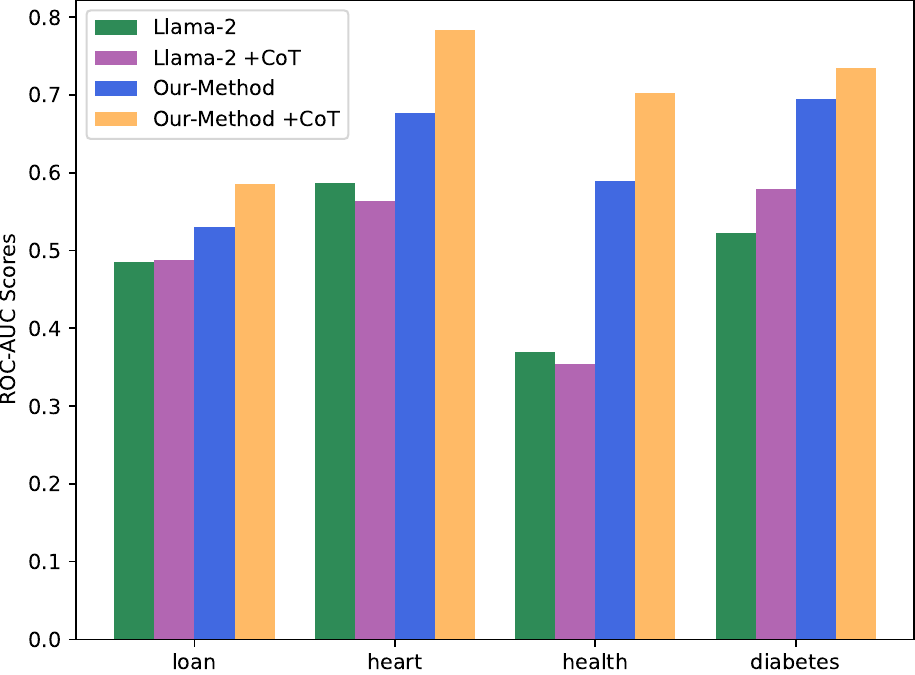}}
\caption{Analysis of predicting target value in the manner of filling in missing values across four tasks. The CoT prompting is also integrated into models. Our method outperforms Llama-2, showing effective reasoning over tabular contents and consistent gains.}
\label{fig/predict_as_fill_in_miss_val}
\end{center}
\end{figure}

\noindent\textbf{Learning Rate Analysis} We evaluate the impact of learning rate (LR) on fine-tuning performance using the credit prediction task, testing LR values of \(2e{-4}\), \(2e{-5}\), \(2e{-6}\), \(2e{-7}\), and \(2e{-8}\). As shown in Figure~\ref{fig/lr_analysis}, small learning rates (smaller than \(2e{-6}\)) yield rapid performance gains in the first four epochs. This observation aligns with~\cite{lima_zhou_2024}, which reports that foundational LLMs can achieve substantial improvements with limited data exposure during instruction tuning. Beyond this early phase, we observe a diminishing improvement trend, indicating that further training yields only marginal performance gains. Additionally, we identify a critical point where the performance reaches a peak, after which a slight decline is observed. We hypothesize that in the early epochs, the model effectively enhances its table reasoning capabilities, but prolonged training at an inappropriate LR may lead to overfitting to the training data.  

Among the evaluated settings, we find that excessively large or small learning rates can hinder optimization, and a learning rate of \(2e{-5}\) consistently outperforms other choices, demonstrating superior performance across multiple epochs. Based on this empirical observation, we select \(2e{-5}\) as the optimal learning rate for our experiments, striking a balance between effective learning and mitigating overfitting.

\noindent\textbf{Finetuning with LoRA} In practical scenarios, fine-tuning large pretrained models on downstream tasks often faces two key challenges: limited dataset size and constrained computational resources. When the available training data is small, full fine-tuning risks overfitting, reducing the model’s generalization ability. Additionally, the high memory and computational demands of updating all model parameters make full fine-tuning impractical on resource-limited hardware. To evaluate the impact of apllying LoRA on fine-tuning performance, we conduct an analysis of different LoRA hyperparameter configurations, specifically varying the LoRA rank (\(r\)) and LoRA scaling factor (\(\alpha\)). The rank \(r\) determines the capacity of the low-rank adaptation, influencing how much flexibility the model has in learning task-specific modifications. The scaling factor \(\alpha\) adjusts the magnitude of LoRA updates relative to the frozen pretrained weights, affecting both convergence behavior and generalization.  

To evaluate the impact of LoRA hyperparameters on fine-tuning performance, we conduct an systematic analysis on a credit prediction task using a dataset of 1,000 examples. Specifically, we vary the $\alpha$ and $r$, assessing their effects on model performance, measured by AUC.
The experimental results, summarized in Figure~\ref{fig/lora_analysis}, reveal distinct performance trends across different LoRA configurations. Higher-rank adaptation leads to the highest AUC scores. However, increasing LoRA rank and scaling factor amplifies training costs. 
Low-rank adaptation with minimal scaling results in suboptimal performance. This suggests that insufficient adaptation capacity hinders the model’s ability to learn effective task-specific reasoning. Based on these results, we recommend ($\alpha$=16, $r$=8) as the trade-off configuration when facing limited fine-tuning examples, as it achieves competitive predictive performance while maintaining efficiency in resource-constrained settings. 

An interesting observation from our analysis is that even with the smallest LoRA configuration ($\alpha$=4,$r$=2), the model maintains modest performance drop against higher-rank settings. This indicates that the LLM, having been pretrained over a massive corpus of tabular data, learns robust table reasoning abilities and effective task transferability, enabling it to adapt to downstream tasks with minimal parameter updates.

\noindent\textbf{Predicting as Imputing Missing Value \& CoT Prompting} 
We explore the feasibility of predicting target values by filling in missing values in the original table. This is achieved by adding a new column that denotes predicted targets for classification tasks, with the model subsequently tasked with predicting these missing values. The results, presented in Figure~\ref{fig/predict_as_fill_in_miss_val}, demonstrate the efficacy of our approach. The notable performance gain of our method over Llama-2 suggests that our model adeptly captures intrinsic relationships within the table and conducts reasoning over the tabular contents before predicting missing values. Additionally, we analyze the impact of combining our models with Chain-of-Thought (CoT) prompting~\cite{wei2022chain}. The original instruction is supplemented with: ``Let's think step by step. You need to first provide the predicted value in the placeholder of \textless missing\_value\_0\textgreater, and then explain your reasons or thoughts.''. This aims to encourage the model to engage in step-by-step reasoning, thereby improving its ability to predict and justify missing values. When combined with CoT prompting, our model consistently demonstrates performance improvements compared to the baseline, indicating its potential to leverage the properties of LLMs to further boost the performance of the trained LLM while excelling in understanding tabular data.

\begin{figure}[!ht]
\begin{center}
\centerline{\includegraphics[width=1.0\columnwidth]{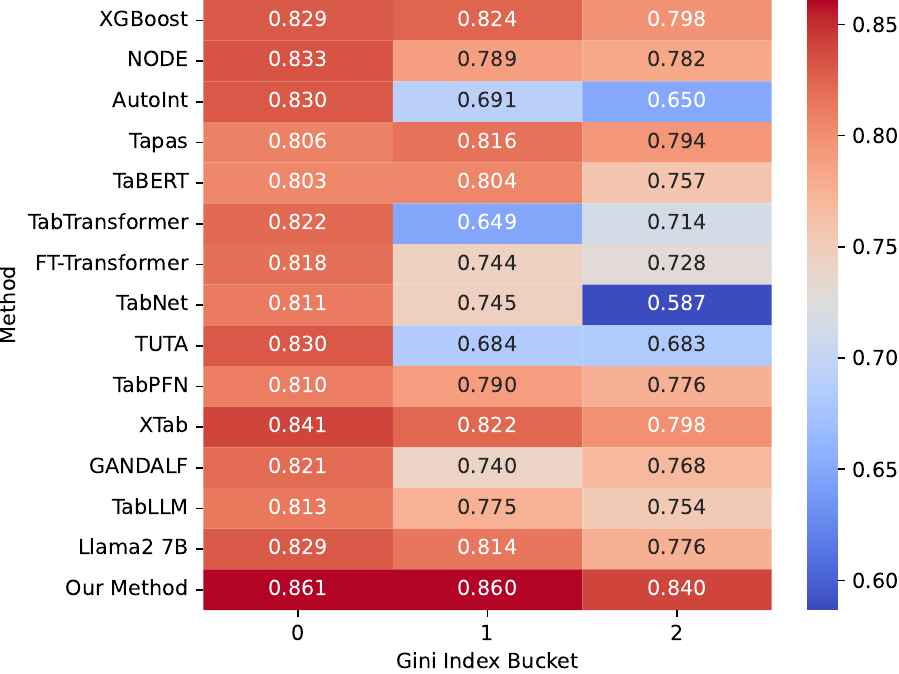}}
\caption{Impact of label imbalance (measured by Gini Index) on model performance. Datasets are grouped by Gini Index, and average ROC-AUC is reported. Our method shows smaller performance decline under high imbalance, highlighting its robustness.}
\label{fig/label_imbalance_analysis}
\end{center}
\end{figure}

\noindent\textbf{Performance Analysis of Label Imbalance}  
This analysis further investigates the influence of imbalanced class distributions on the performance of the proposed model by employing the Gini Index as a metric for quantifying class distribution inequality. The datasets are divided into three distinct categories according to their Gini Index values, and the average ROC-AUC score is calculated for each method within these categories. The findings, illustrated in Figure~\ref{fig/label_imbalance_analysis}, indicate that while label imbalance poses a substantial challenge for competing algorithms, the method introduced in this work demonstrates a comparatively modest decline in performance under such conditions. This observation underscores the enhanced robustness and effectiveness of the proposed approach, which, due to its extensive pretraining on tabular data, is better equipped to handle the complexities associated with uneven class distributions.

\subsection{Pretrain Broader Spectrum of LLMs}
\label{subsec_pretrain_more_llms}
In our previous experiments, we pre-trained Llama-2 with massive tabular data and observed clear improvements in its performance on table-related predictive tasks. Motivated by these promising results, we seek to investigate whether this pretraining approach could be extended to other large language models (LLMs). Specifically, we aim to test the transferability of continued pretraining on tables across various open-sourced LLMs. To this end, we select three other LLMs (Llama-3.1 8B, Qwen-2.5 7B, and DeepSeek-R1-distilled Llama-3.1 8B) to evaluate whether the performance improvements observed with Llama-2 could be replicated or even enhanced by pretraining these models with tabular data.

\begin{figure}[!ht]
\begin{center}
\centerline{\includegraphics[width=1.0\columnwidth]{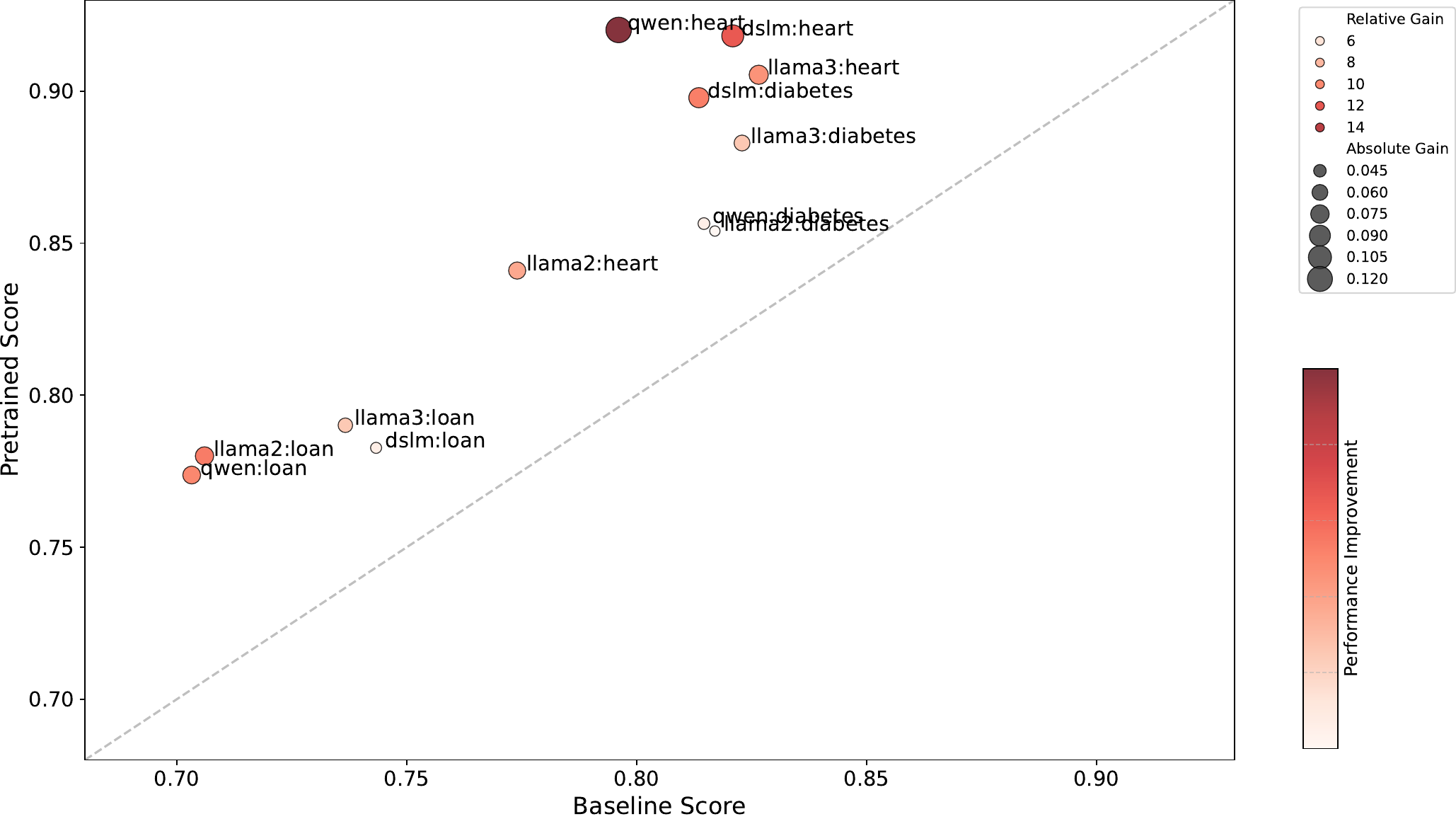}}
\caption{Performance improvement of various models on table-related predictive tasks. The x-axis represents the baseline scores, and the y-axis represents the pretrained scores. The bubble size corresponds to the performance improvement magnitude, with larger bubbles indicating higher performance gain after pretraining. Darker colors also indicate higher improvements. The results show substantial performance gains for pretrained models (denoted with ``pretrained-'') across different LLMs and datasets. The ``dslm'' here denotes DeepSeek-R1-distilled Llama.}
\label{fig/other_llm_transfer}
\end{center}
\end{figure}

In this experiment, we finetune the baseline models (without pretraining) and their respective pretrained counterparts on three tabular datasets: ``heart'', ``loan'', and ``diabetes''. The pretrained models are obtained with the same approach that proved successful for Llama-2. After finetuning, performance is compared between the baseline and pretrained models, with the results visualized using a bubble chart in Figure~\ref{fig/other_llm_transfer}. The x-axis represents the baseline score, the y-axis represents the pretrained score, and the size and color intensity of the bubbles indicate performance improvement levels.

As we can see from Figure~\ref{fig/other_llm_transfer}, all the bubbles are positioned above the diagonal line, further highlighting that the ``pretrained-'' models achieve better scores compared to their baseline counterparts, with varying degrees of improvement. The experimental results confirm that continued pretraining with tabular data leads to consistent and significant performance improvements across all models and datasets. The pretrained models outperformed their baseline versions across the board. Notably, pretrained-QWen showed a considerable improvement on the ``heart'' dataset, where the score increased from 0.79 to 0.92, demonstrating a substantial gain of around 13\%. Similar performance boosts were observed for pretrained models on the ``loan'' and ``diabetes'' datasets, indicating that the pretraining approach is effective across multiple LLMs.

These results validate our hypothesis that the benefits of continued pretraining with massive tabular data are transferable to other open-sourced LLM architectures. The consistent performance improvements across different models provide strong evidence that this pretraining strategy enhances the ability of LLMs to perform well on table-related predictive tasks, further expanding the potential application of LLMs in this domain.

\subsection{Case Demonstration}
\label{subsec_case_study}
To verify the effectiveness of our approach across different predictive tasks, we present case studies for three fundamental problem settings: missing value prediction, classification, and regression. For each task, we employ a unified prompt-based framework that encodes task-specific instructions and tabular content into a structured input format for the model.

\begin{figure}[!ht]
\begin{center}
\centerline{\includegraphics[width=0.9\columnwidth]{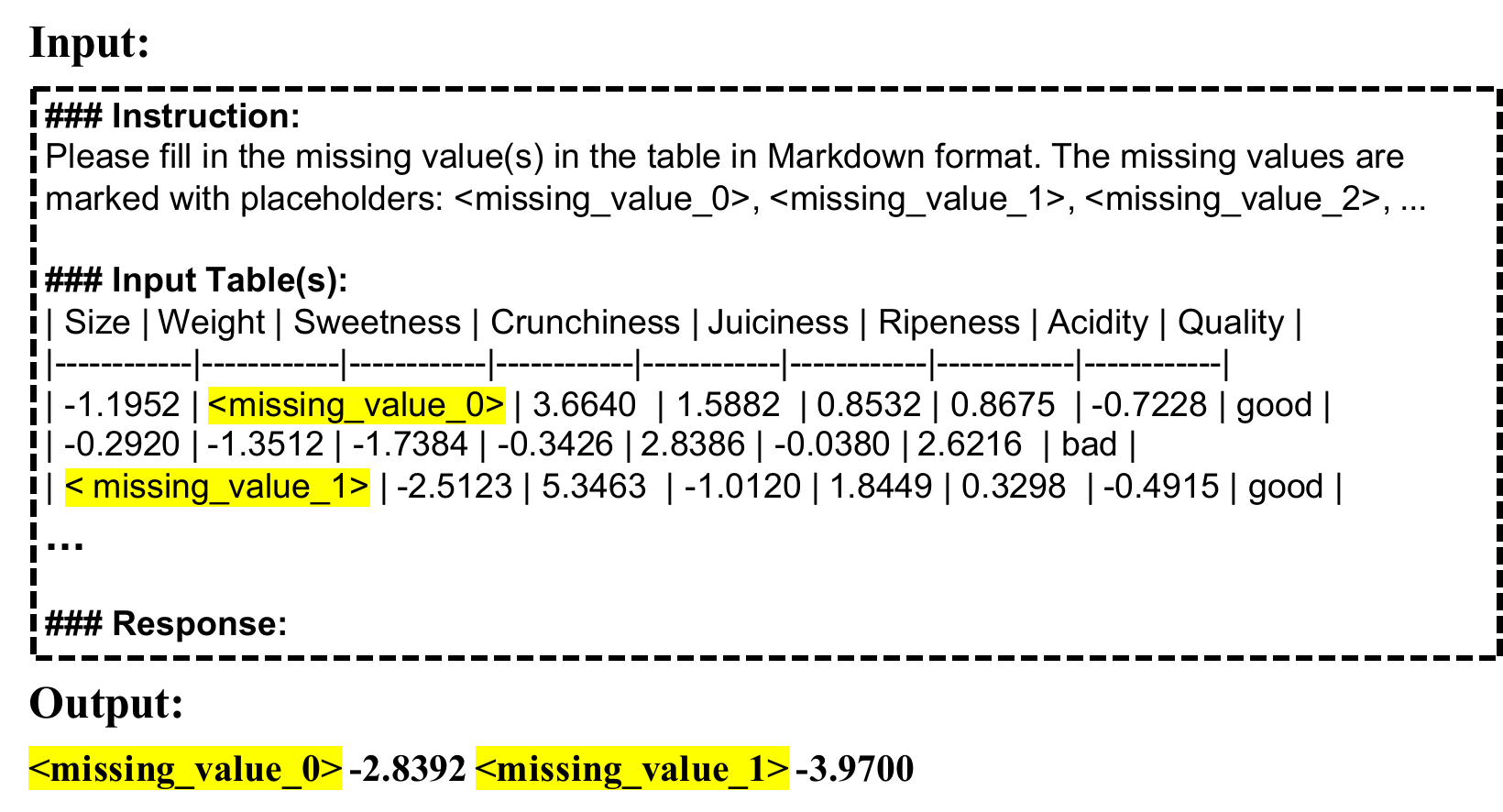}}
\caption{Prompt for the task of missing value prediction. The concealed cells are marked with specific sentinel tokens. The model is expected to predict all masked contents with corresponding sentinel words. The sentinel tokens indicating missing values are highlighted for better demonstration.}
\label{fig/prompt_demo_mask}
\end{center}
\end{figure}

\textbf{Missing Value Prediction} 
Figure~\ref{fig/prompt_demo_mask} illustrates an example prompt for the missing value prediction task. In this setting, selected table cells are intentionally masked and replaced with sentinel tokens, which act as placeholders for missing values. The model is provided with a structured prompt consisting of task instructions and the partially observed table, guiding it to infer and generate the missing content at the designated sentinel positions. This design allows the model to capture patterns within tabular data by leveraging both explicit and implicit relationships among attributes. The highlighted sentinel tokens in Figure~\ref{fig/prompt_demo_mask} further aid in visualizing the missing value imputation process.

\begin{figure}[htb]
\begin{center}
\centerline{\includegraphics[width=0.9\columnwidth]{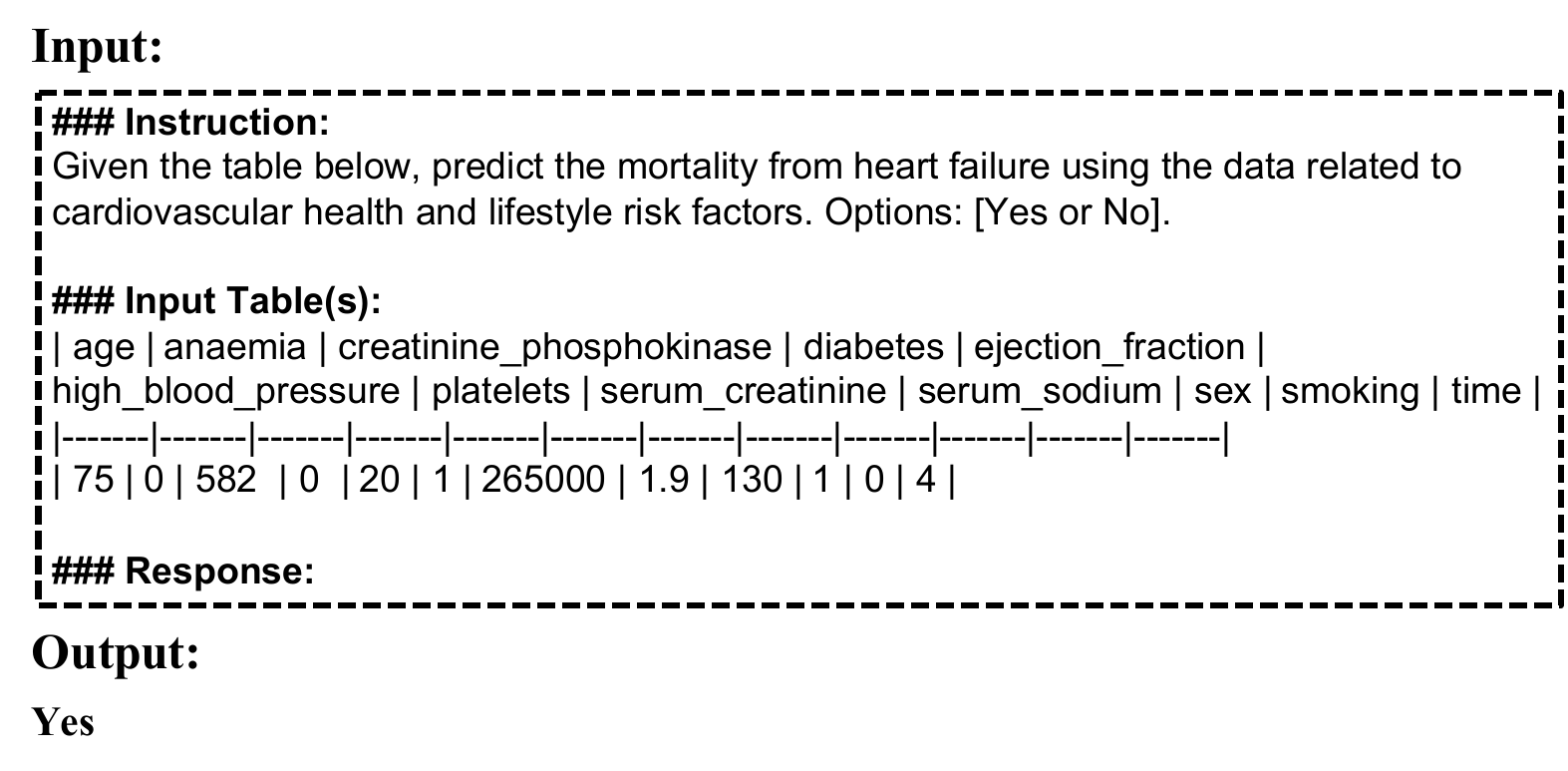}}
\vskip -1ex
\caption{Prompt for the classification task. The model is asked to predict the target class according to the given instruction and tabular content. In this demonstration case, the model is required to learn to predict the mortality from the give table.}
\label{fig/prompt_demo_cls}
\end{center}
\end{figure}

\textbf{Classification} 
In addition to missing value prediction, Figure~\ref{fig/prompt_demo_cls} illustrates the prompt structure for a classification task. The input consists of a task-specific instruction paired with the tabular content, formatted according to our unified prompt template. The instruction concisely describes the classification task and specifies the set of possible output categories. In the example shown, the model is asked to predict patient mortality due to heart failure based on relevant clinical attributes. This structured prompt provides clear guidance on the classification objective, instructing the model to generate the correct target label among ``Yes'' or ``No''.

\begin{figure}[t!]
\begin{center}
\centerline{\includegraphics[width=0.88\columnwidth]{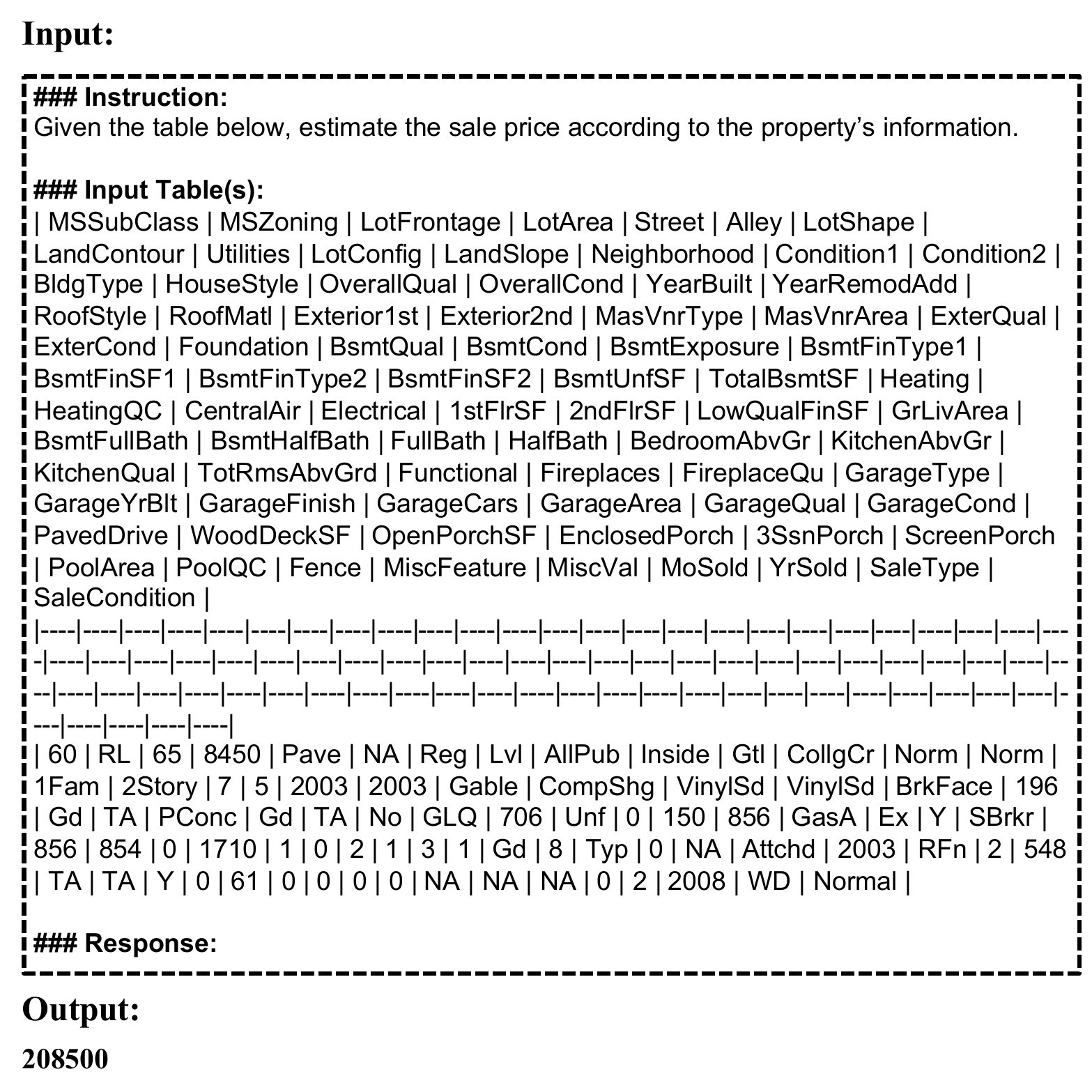}}
\vskip -1ex
\caption{Prompt for the regression task. The model is asked to predict the target value according to the given instruction and tabular content. In this demonstrated case, the model is required to learn to predict the sale price of a house.}
\label{fig/prompt_demo_reg}
\end{center}
\end{figure}

\textbf{Regression} 
Figure~\ref{fig/prompt_demo_reg} illustrates the prompt design for a regression task. As in the classification setup, the model input consists of a task-specific instruction specifying the prediction objective, together with the structured tabular data. Unlike classification, where the target is a discrete label, the model here is trained to predict a continuous numerical value. In the example shown, the model is tasked with estimating the sale price of a house based on descriptive attributes such as location, size, and features. As mentioned earlier, it is also feasible to introduce an additional output head to directly predict the regression target value. The prompt-based design enables the model to effectively learn regression patterns and generalize to new, unseen data distributions.

\section{Conclusion}
\label{conclusion_sect} 
This work addresses the gap between large language models (LLMs) and their application to predictive tabular tasks, a critical yet underexplored area in LLM research. By pretraining existing LLMs on a curated dataset comprising approximately 13 billion examples sourced from Kaggle and other tabular-focused repositories, we demonstrate significant improvements in key predictive tasks, including classification, regression, and missing value prediction. The pretrained models outperform baseline methods, including GPT-4, particularly on tasks involving missing value imputation. 
We further validate the generalizability of our approach by applying it to multiple open-source LLMs, showing that the benefits of our method are transferable across different architectures. Beyond these improvements, our approach proves effective in few-shot prediction and long-context learning scenarios, highlighting the versatility and robustness of LLMs when adapted to structured tabular data. 
Overall, our results underscore the potential of well-pretrained LLMs to enhance performance on tabular tasks and provide a strong foundation for future research in this domain. This work lays the groundwork for advancing the application of LLMs in data science, with a particular focus on refining analytical and predictive modeling capabilities for table-related tasks. The demonstrated generalizability of our method opens promising avenues for leveraging LLMs across a broader spectrum of structured data challenges.

\section*{Limitations}
This work focuses exclusively on tabular tasks composed of numerical and textual values, deliberately excluding other data modalities such as images, audio, and video. This limitation may reduce the applicability of our model in scenarios that require integration of multimedia data, potentially affecting its generalizability. Furthermore, our approach differs from conventional pretraining methodologies, which are primarily designed for text-generation tasks. By concentrating on predictive tasks relevant to data science, our pretraining objectives are specifically restricted to Mask-Then-Predict and Multi-Task training. While this specialization benefits targeted applications, it limits the model's ability to leverage a broader set of pretraining objectives, potentially constraining its adaptability to other text-generation tasks outside the scoped domains. 
In addition, our model has only been evaluated on English data, which restricts its effectiveness and reliability in multilingual contexts and may lead to errors or misinterpretations when applied to other languages. Moreover, the model currently does not support nested tables, limiting its coverage and the complexity of table structures it can effectively handle.

\vfill

\bibliographystyle{IEEEtran}
\bibliography{insttab}

@inproceedings{wang2024chain,
  title={Chain-of-Table: Evolving Tables in the Reasoning Chain for Table Understanding},
  author={Wang, Zilong and Zhang, Hao and Li, Chun-Liang and Eisenschlos, Julian Martin and Perot, Vincent and Wang, Zifeng and Miculicich, Lesly and Fujii, Yasuhisa and Shang, Jingbo and Lee, Chen-Yu and others},
  booktitle={The Twelfth International Conference on Learning Representations}
}

@inproceedings{zhang2023tablellama,
  title={Tablellama: Towards open large generalist models for tables},
  author={Zhang, Tianshu and Yue, Xiang and Li, Yifei and Sun, Huan},
  booktitle={Proceedings of the 2024 Conference of the North American Chapter of the Association for Computational Linguistics: Human Language Technologies (Volume 1: Long Papers)},
  pages={6024--6044},
  year={2024}
}

@inproceedings{chen2016xgboost,
  title={Xgboost: A scalable tree boosting system},
  author={Chen, Tianqi and Guestrin, Carlos},
  booktitle={Proceedings of the 22nd acm sigkdd international conference on knowledge discovery and data mining},
  pages={785--794},
  year={2016}
}

@inproceedings{yin2020tabert,
  title={TaBERT: Pretraining for joint understanding of textual and tabular data},
  author={Yin, Pengcheng and Neubig, Graham and Yih, Wen-tau and Riedel, Sebastian},
  booktitle={Proceedings of the 58th annual meeting of the association for computational linguistics},
  pages={8413--8426},
  year={2020}
}

@inproceedings{wang2021tuta,
  title={TUTA: tree-based transformers for generally structured table pre-training},
  author={Wang, Zhiruo and Dong, Haoyu and Jia, Ran and Li, Jia and Fu, Zhiyi and Han, Shi and Zhang, Dongmei},
  booktitle={Proceedings of the 27th ACM SIGKDD Conference on Knowledge Discovery \& Data Mining},
  pages={1780--1790},
  year={2021}
}

@article{wang2022transtab,
  title={Transtab: Learning transferable tabular transformers across tables},
  author={Wang, Zifeng and Sun, Jimeng},
  journal={Advances in Neural Information Processing Systems},
  volume={35},
  pages={2902--2915},
  year={2022}
}

@inproceedings{devlin2018bert,
  title={Bert: Pre-training of deep bidirectional transformers for language understanding},
  author={Devlin, Jacob and Chang, Ming-Wei and Lee, Kenton and Toutanova, Kristina},
  booktitle={Proceedings of the 2019 conference of the North American chapter of the association for computational linguistics: human language technologies, volume 1 (long and short papers)},
  pages={4171--4186},
  year={2019}
}

@article{liu2022ptab,
  title={PTab: Using the Pre-trained Language Model for Modeling Tabular Data},
  author={Liu, Guang and Yang, Jie and Wu, Ledell},
  journal={arXiv preprint arXiv:2209.08060},
  year={2022}
}

@inproceedings{yin20tabert,
  title={TaBERT: Pretraining for joint understanding of textual and tabular data},
  author={Yin, Pengcheng and Neubig, Graham and Yih, Wen-tau and Riedel, Sebastian},
  booktitle={Proceedings of the 58th annual meeting of the association for computational linguistics},
  pages={8413--8426},
  year={2020}
}

@article{huang2020tabtransformer,
  title={Tabtransformer: Tabular data modeling using contextual embeddings},
  author={Huang, Xin and Khetan, Ashish and Cvitkovic, Milan and Karnin, Zohar},
  journal={arXiv preprint arXiv:2012.06678},
  year={2020}
}

@article{gorishniy2022embeddings,
  title={On embeddings for numerical features in tabular deep learning},
  author={Gorishniy, Yury and Rubachev, Ivan and Babenko, Artem},
  journal={Advances in Neural Information Processing Systems},
  volume={35},
  pages={24991--25004},
  year={2022}
}

@inproceedings{herzig2020tapas,
  title={TaPas: Weakly supervised table parsing via pre-training},
  author={Herzig, Jonathan and Nowak, Pawel Krzysztof and M{\"u}ller, Thomas and Piccinno, Francesco and Eisenschlos, Julian},
  booktitle={Proceedings of the 58th annual meeting of the association for computational linguistics},
  pages={4320--4333},
  year={2020}
}

@inproceedings{hollmann2022tabpfn,
title={Tab{PFN}: A Transformer That Solves Small Tabular Classification Problems in a Second},
author={Noah Hollmann and Samuel M{\"u}ller and Katharina Eggensperger and Frank Hutter},
booktitle={The Eleventh International Conference on Learning Representations },
year={2023},
url={https://openreview.net/forum?id=cp5PvcI6w8_}
}

@article{gorishniy2021revisiting,
  title={Revisiting deep learning models for tabular data},
  author={Gorishniy, Yury and Rubachev, Ivan and Khrulkov, Valentin and Babenko, Artem},
  journal={Advances in Neural Information Processing Systems},
  volume={34},
  pages={18932--18943},
  year={2021}
}

@article{joseph2023gandalf,
  title={GANDALF: gated adaptive network for deep automated learning of features},
  author={Joseph, Manu and Raj, Harsh},
  journal={arXiv preprint arXiv:2207.08548},
  year={2022}
}

@inproceedings{popov2019neural,
title={Neural Oblivious Decision Ensembles for Deep Learning on Tabular Data},
author={Sergei Popov and Stanislav Morozov and Artem Babenko},
booktitle={International Conference on Learning Representations},
year={2020},
url={https://openreview.net/forum?id=r1eiu2VtwH}
}

@inproceedings{vaswani2017attention,
 author = {Vaswani, Ashish and Shazeer, Noam and Parmar, Niki and Uszkoreit, Jakob and Jones, Llion and Gomez, Aidan N and Kaiser, \L ukasz and Polosukhin, Illia},
 booktitle = {Advances in Neural Information Processing Systems},
 editor = {I. Guyon and U. Von Luxburg and S. Bengio and H. Wallach and R. Fergus and S. Vishwanathan and R. Garnett},
 pages = {},
 publisher = {Curran Associates, Inc.},
 title = {Attention is All you Need},
 url = {https://proceedings.neurips.cc/paper_files/paper/2017/file/3f5ee243547dee91fbd053c1c4a845aa-Paper.pdf},
 volume = {30},
 year = {2017}
}

@article{grinsztajn2022tree,
  title={Why do tree-based models still outperform deep learning on typical tabular data?},
  author={Grinsztajn, L{\'e}o and Oyallon, Edouard and Varoquaux, Ga{\"e}l},
  journal={Advances in Neural Information Processing Systems},
  volume={35},
  pages={507--520},
  year={2022}
}

@article{brown2020language,
  title={Language models are few-shot learners},
  author={Brown, Tom and Mann, Benjamin and Ryder, Nick and Subbiah, Melanie and Kaplan, Jared D and Dhariwal, Prafulla and Neelakantan, Arvind and Shyam, Pranav and Sastry, Girish and Askell, Amanda and others},
  journal={Advances in neural information processing systems},
  volume={33},
  pages={1877--1901},
  year={2020}
}

@article{collins2022fedavg,
  title={Fedavg with fine tuning: Local updates lead to representation learning},
  author={Collins, Liam and Hassani, Hamed and Mokhtari, Aryan and Shakkottai, Sanjay},
  journal={Advances in Neural Information Processing Systems},
  volume={35},
  pages={10572--10586},
  year={2022}
}

@inproceedings{hegselmann2023tabllm,
  title={Tabllm: Few-shot classification of tabular data with large language models},
  author={Hegselmann, Stefan and Buendia, Alejandro and Lang, Hunter and Agrawal, Monica and Jiang, Xiaoyi and Sontag, David},
  booktitle={International Conference on Artificial Intelligence and Statistics},
  pages={5549--5581},
  year={2023},
  organization={PMLR}
}

@inproceedings{gu2022pasta,
  title={PASTA: table-operations aware fact verification via sentence-table cloze pre-training},
  author={Gu, Zihui and Fan, Ju and Tang, Nan and Nakov, Preslav and Zhao, Xiaoman and Du, Xiaoyong},
  booktitle={Proceedings of the 2022 conference on empirical methods in natural language processing},
  pages={4971--4983},
  year={2022}
}

@article{nan2022fetaqa,
  title={FeTaQA: Free-form table question answering},
  author={Nan, Linyong and Hsieh, Chiachun and Mao, Ziming and Lin, Xi Victoria and Verma, Neha and Zhang, Rui and Kry{\'s}ci{\'n}ski, Wojciech and Schoelkopf, Hailey and Kong, Riley and Tang, Xiangru and others},
  journal={Transactions of the Association for Computational Linguistics},
  volume={10},
  pages={35--49},
  year={2022},
  publisher={MIT Press One Broadway, 12th Floor, Cambridge, Massachusetts 02142, USA~…}
}

@inproceedings{xiong2023effective,
  title={Effective long-context scaling of foundation models},
  author={Xiong, Wenhan and Liu, Jingyu and Molybog, Igor and Zhang, Hejia and Bhargava, Prajjwal and Hou, Rui and Martin, Louis and Rungta, Rashi and Sankararaman, Karthik Abinav and Oguz, Barlas and others},
  booktitle={Proceedings of the 2024 Conference of the North American Chapter of the Association for Computational Linguistics: Human Language Technologies (Volume 1: Long Papers)},
  pages={4643--4663},
  year={2024}
}

@inproceedings{fu2024data,
title={Data Engineering for Scaling Language Models to 128K Context},
author={Yao Fu and Rameswar Panda and Xinyao Niu and Xiang Yue and Hannaneh Hajishirzi and Yoon Kim and Hao Peng},
booktitle={Forty-first International Conference on Machine Learning},
year={2024},
url={https://openreview.net/forum?id=TaAqeo7lUh}
}

@inproceedings{zhu2023xtab,
  title={XTab: cross-table pretraining for tabular transformers},
  author={Zhu, Bingzhao and Shi, Xingjian and Erickson, Nick and Li, Mu and Karypis, George and Shoaran, Mahsa},
  booktitle={Proceedings of the 40th International Conference on Machine Learning},
  pages={43181--43204},
  year={2023}
}

@inproceedings{lin2004rouge,
    title = "{ROUGE}: A Package for Automatic Evaluation of Summaries",
    author = "Lin, Chin-Yew",
    booktitle = "Text Summarization Branches Out",
    month = jul,
    year = "2004",
    address = "Barcelona, Spain",
    publisher = "Association for Computational Linguistics",
    url = "https://aclanthology.org/W04-1013/",
    pages = "74--81"
}

@article{li2023table,
  title={Table-gpt: Table-tuned gpt for diverse table tasks},
  author={Li, Peng and He, Yeye and Yashar, Dror and Cui, Weiwei and Ge, Song and Zhang, Haidong and Fainman, Danielle Rifinski and Zhang, Dongmei and Chaudhuri, Surajit},
  journal={arXiv preprint arXiv:2310.09263},
  year={2023}
}

@inproceedings{zhao2023investigating,
  title={Investigating Table-to-Text Generation Capabilities of Large Language Models in Real-World Information Seeking Scenarios},
  author={Zhao, Yilun and Zhang, Haowei and Si, Shengyun and Nan, Linyong and Tang, Xiangru and Cohan, Arman},
  booktitle={Proceedings of the 2023 Conference on Empirical Methods in Natural Language Processing: Industry Track},
  pages={160--175},
  year={2023}
}

@inproceedings{shin2023arxiveri,
title={arXiVeri: Automatic table verification with {GPT}},
author={Gyungin Shin and Weidi Xie and Samuel Albanie},
booktitle={NeurIPS 2023 AI for Science Workshop},
year={2023},
url={https://openreview.net/forum?id=v6aJLdXsZx}
}

@article{wei2022chain,
  title={Chain-of-thought prompting elicits reasoning in large language models},
  author={Wei, Jason and Wang, Xuezhi and Schuurmans, Dale and Bosma, Maarten and Xia, Fei and Chi, Ed and Le, Quoc V and Zhou, Denny and others},
  journal={Advances in neural information processing systems},
  volume={35},
  pages={24824--24837},
  year={2022}
}

@article{slack2023tablet,
  title={Tablet: Learning from instructions for tabular data},
  author={Slack, Dylan and Singh, Sameer},
  journal={arXiv preprint arXiv:2304.13188},
  year={2023}
}

@inproceedings{yang2024unitabe,
title={UniTabE: A Universal Pretraining Protocol for Tabular Foundation  Model in Data Science},
author={Yazheng Yang and Yuqi Wang and Guang Liu and Ledell Wu and Qi Liu},
booktitle={The Twelfth International Conference on Learning Representations},
year={2024},
url={https://openreview.net/forum?id=6LLho5X6xV}
}

@inproceedings{gong2020tablegpt,
  title={Tablegpt: Few-shot table-to-text generation with table structure reconstruction and content matching},
  author={Gong, Heng and Sun, Yawei and Feng, Xiaocheng and Qin, Bing and Bi, Wei and Liu, Xiaojiang and Liu, Ting},
  booktitle={Proceedings of the 28th International Conference on Computational Linguistics},
  pages={1978--1988},
  year={2020}
}

@inproceedings{cheng2021hitab,
  title={Hitab: A hierarchical table dataset for question answering and natural language generation},
  author={Cheng, Zhoujun and Dong, Haoyu and Wang, Zhiruo and Jia, Ran and Guo, Jiaqi and Gao, Yan and Han, Shi and Lou, Jian-Guang and Zhang, Dongmei},
  booktitle={Proceedings of the 60th Annual Meeting of the Association for Computational Linguistics (Volume 1: Long Papers)},
  pages={1094--1110},
  year={2022}
}

@article{deng2022turl,
  title={Turl: Table understanding through representation learning},
  author={Deng, Xiang and Sun, Huan and Lees, Alyssa and Wu, You and Yu, Cong},
  journal={ACM SIGMOD Record},
  volume={51},
  number={1},
  pages={33--40},
  year={2022},
  publisher={ACM New York, NY, USA}
}

@inproceedings{parikh2020totto,
  title={ToTTo: A controlled table-to-text generation dataset},
  author={Parikh, Ankur and Wang, Xuezhi and Gehrmann, Sebastian and Faruqui, Manaal and Dhingra, Bhuwan and Yang, Diyi and Das, Dipanjan},
  booktitle={Proceedings of the 2020 Conference on Empirical Methods in Natural Language Processing (EMNLP)},
  pages={1173--1186},
  year={2020}
}

@inproceedings{lima_zhou_2024,
 author = {Zhou, Chunting and Liu, Pengfei and Xu, Puxin and Iyer, Srinivasan and Sun, Jiao and Mao, Yuning and Ma, Xuezhe and Efrat, Avia and Yu, Ping and YU, LILI and Zhang, Susan and Ghosh, Gargi and Lewis, Mike and Zettlemoyer, Luke and Levy, Omer},
 booktitle = {Advances in Neural Information Processing Systems},
 editor = {A. Oh and T. Naumann and A. Globerson and K. Saenko and M. Hardt and S. Levine},
 pages = {55006--55021},
 publisher = {Curran Associates, Inc.},
 title = {LIMA: Less Is More for Alignment},
 url = {https://proceedings.neurips.cc/paper_files/paper/2023/file/ac662d74829e4407ce1d126477f4a03a-Paper-Conference.pdf},
 volume = {36},
 year = {2023}
}

@inproceedings{nararatwong-etal-2022-enhancing,
    title = "Enhancing Financial Table and Text Question Answering with Tabular Graph and Numerical Reasoning",
    author = "Nararatwong, Rungsiman  and
      Kertkeidkachorn, Natthawut  and
      Ichise, Ryutaro",
    editor = "He, Yulan  and
      Ji, Heng  and
      Li, Sujian  and
      Liu, Yang  and
      Chang, Chua-Hui",
    booktitle = "Proceedings of the 2nd Conference of the Asia-Pacific Chapter of the Association for Computational Linguistics and the 12th International Joint Conference on Natural Language Processing (Volume 1: Long Papers)",
    month = nov,
    year = "2022",
    address = "Online only",
    publisher = "Association for Computational Linguistics",
    url = "https://aclanthology.org/2022.aacl-main.72/",
    doi = "10.18653/v1/2022.aacl-main.72",
    pages = "991--1000",
}

@inproceedings{he-etal-2023-anameta,
    title = "{A}na{M}eta: A Table Understanding Dataset of Field Metadata Knowledge Shared by Multi-dimensional Data Analysis Tasks",
    author = "He, Xinyi  and
      Zhou, Mengyu  and
      Zhou, Mingjie  and
      Xu, Jialiang  and
      Lv, Xiao  and
      Li, Tianle  and
      Shao, Yijia  and
      Han, Shi  and
      Yuan, Zejian  and
      Zhang, Dongmei",
    editor = "Rogers, Anna  and
      Boyd-Graber, Jordan  and
      Okazaki, Naoaki",
    booktitle = "Findings of the Association for Computational Linguistics: ACL 2023",
    month = jul,
    year = "2023",
    address = "Toronto, Canada",
    publisher = "Association for Computational Linguistics",
    url = "https://aclanthology.org/2023.findings-acl.604/",
    doi = "10.18653/v1/2023.findings-acl.604",
    pages = "9471--9492",
}

@inproceedings{zhao-etal-2023-qtsumm,
    title = "{QTS}umm: Query-Focused Summarization over Tabular Data",
    author = "Zhao, Yilun  and
      Qi, Zhenting  and
      Nan, Linyong  and
      Mi, Boyu  and
      Liu, Yixin  and
      Zou, Weijin  and
      Han, Simeng  and
      Chen, Ruizhe  and
      Tang, Xiangru  and
      Xu, Yumo  and
      Radev, Dragomir  and
      Cohan, Arman",
    editor = "Bouamor, Houda  and
      Pino, Juan  and
      Bali, Kalika",
    booktitle = "Proceedings of the 2023 Conference on Empirical Methods in Natural Language Processing",
    month = dec,
    year = "2023",
    address = "Singapore",
    publisher = "Association for Computational Linguistics",
    url = "https://aclanthology.org/2023.emnlp-main.74/",
    doi = "10.18653/v1/2023.emnlp-main.74",
    pages = "1157--1172",
}

@inproceedings{zhang-etal-2023-generative,
    title = "Generative Table Pre-training Empowers Models for Tabular Prediction",
    author = "Zhang, Tianping  and
      Wang, Shaowen  and
      Yan, Shuicheng  and
      Jian, Li  and
      Liu, Qian",
    editor = "Bouamor, Houda  and
      Pino, Juan  and
      Bali, Kalika",
    booktitle = "Proceedings of the 2023 Conference on Empirical Methods in Natural Language Processing",
    month = dec,
    year = "2023",
    address = "Singapore",
    publisher = "Association for Computational Linguistics",
    url = "https://aclanthology.org/2023.emnlp-main.917/",
    doi = "10.18653/v1/2023.emnlp-main.917",
    pages = "14836--14854",
}

@inproceedings{yang2022tableformer,
  title={TableFormer: Robust transformer modeling for table-text encoding},
  author={Yang, Jingfeng and Gupta, Aditya and Upadhyay, Shyam and He, Luheng and Goel, Rahul and Paul, Shachi},
  booktitle={Proceedings of the 60th Annual Meeting of the Association for Computational Linguistics (Volume 1: Long Papers)},
  pages={528--537},
  year={2022}
}

\section{Biography Section}

\begin{IEEEbiography}[{\includegraphics
[width=1in,height=1.25in,clip,
keepaspectratio]{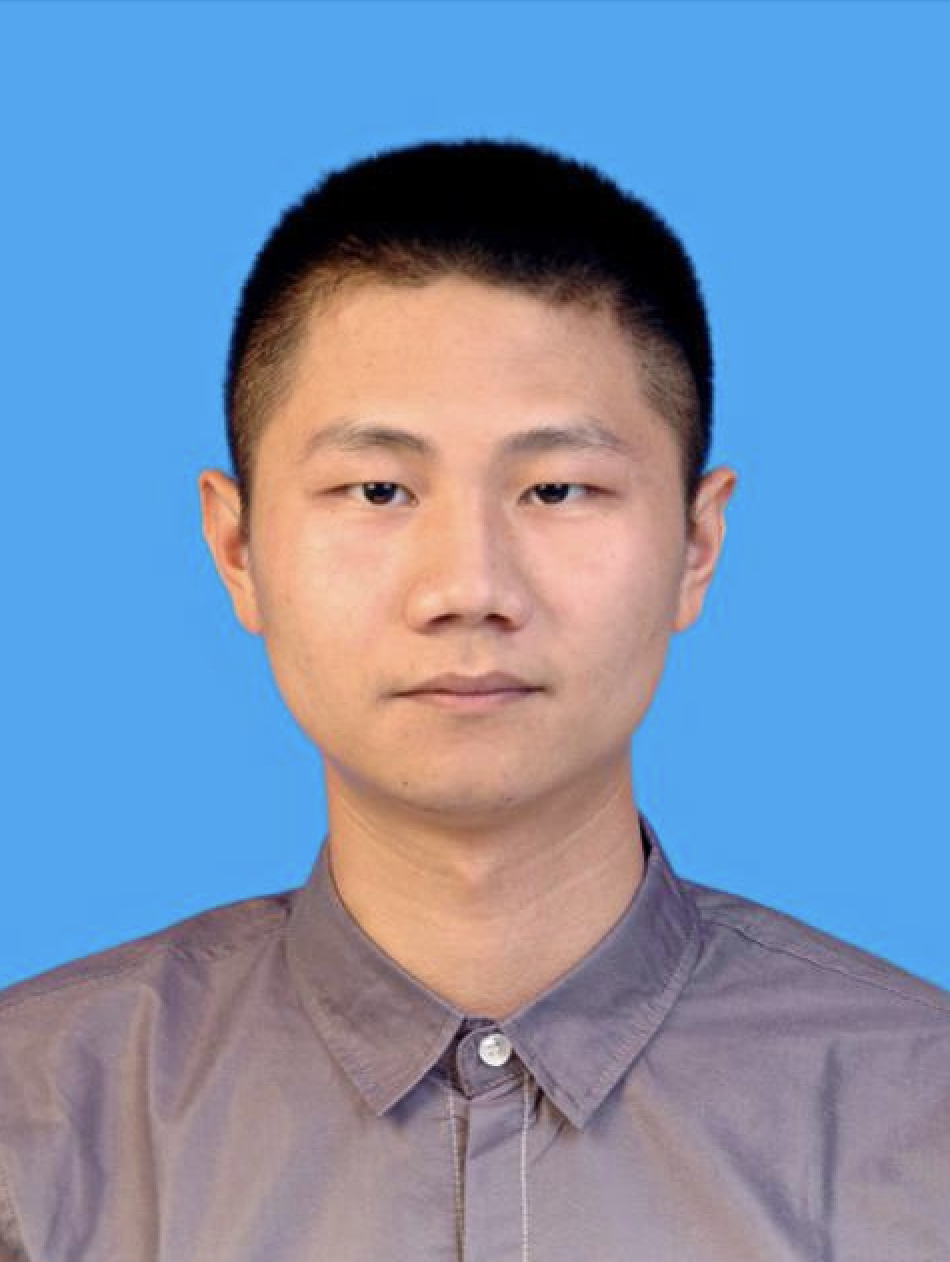}}]
{Yazheng Yang} is a PhD student in the School of Computing and Data Science, the University of Hong Kong. His research interests lie at the intersection of natural language processing and artificial intelligence, with a focus on pre-training large language models across different modalities, particularly tables and speech data, to support practical applications that automate tasks and enhance human productivity. His recent work also explores audio tokenization and the pre-training of speech language models.
\end{IEEEbiography}

\begin{IEEEbiography}[{\includegraphics
[width=1in,height=1.25in,clip,
keepaspectratio]{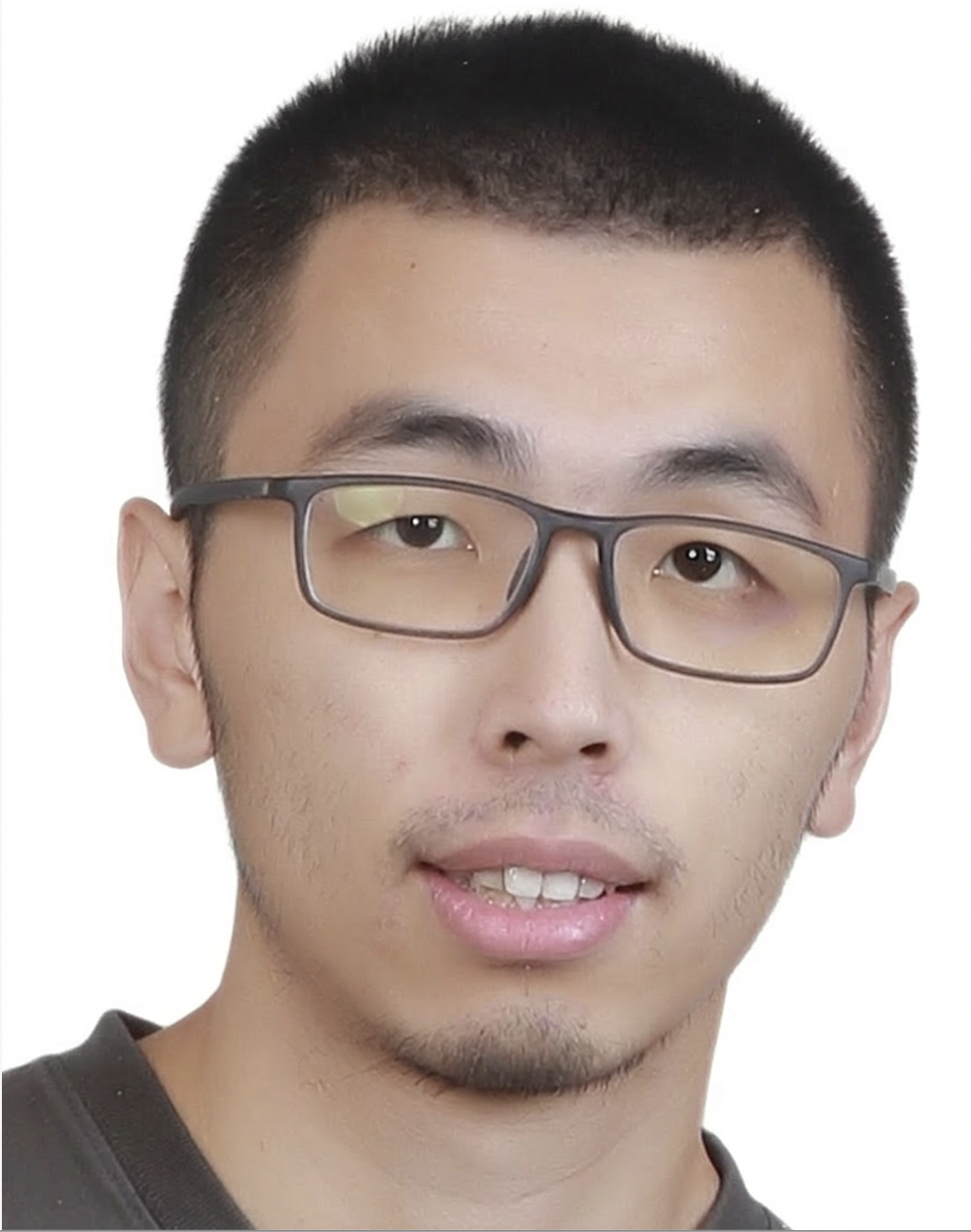}}]
{Yuqi Wang} is a PhD student in School of Computing and Data Science at the University of Hong Kong, supervised by Professor Qi Liu. His research focuses on large language models, multimodal learning, and tabular machine learning. His work has appeared at venues such as ICLR and ACL.
\end{IEEEbiography}

\begin{IEEEbiography}[{\includegraphics
[width=1in,height=1.25in,clip,
keepaspectratio]{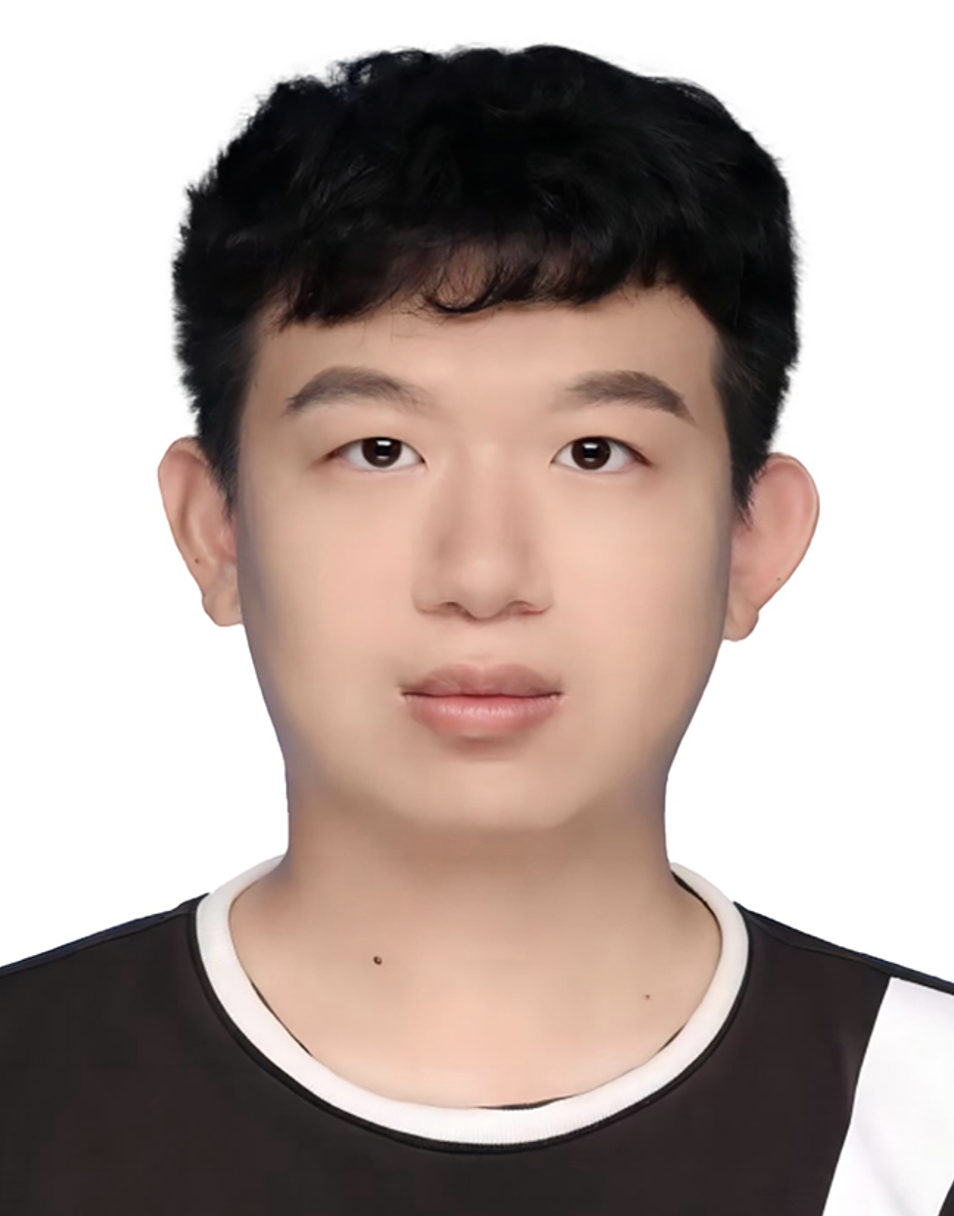}}]
{Yaxuan Li} received the B.Eng. degree in computer science and technology from Harbin Institute of Technology, Shenzhen, in 2024. His research interests include natural language processing and AI-generated content.
\end{IEEEbiography}

\begin{IEEEbiography}[{\includegraphics
[width=1in,height=1.25in,clip,
keepaspectratio]{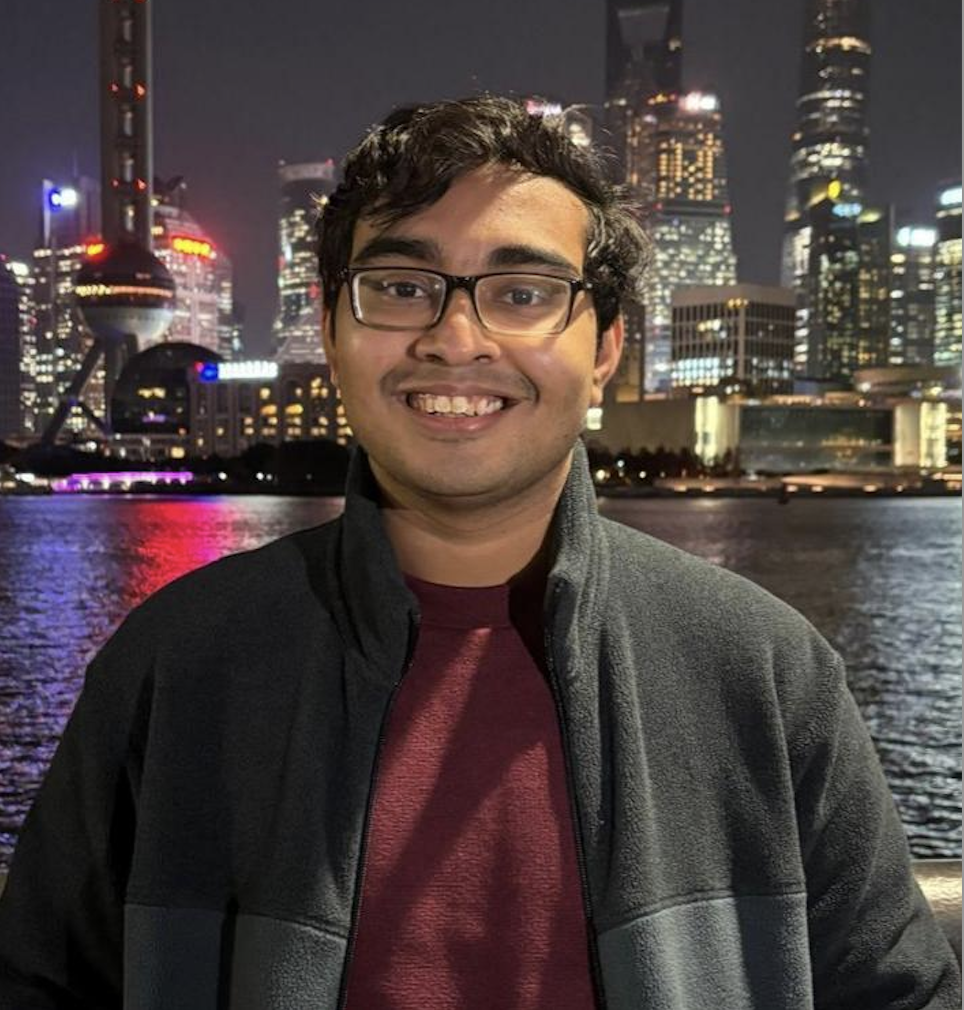}}]
{Sankalok Sen} graduated with a major in Computer Science and a minor in Statistics from The University of Hong Kong in 2023. His research interests lie in building AI systems for real world use cases, and focussing on model optimizations, reproducibility and faithfulness of AI systems.
\end{IEEEbiography}

\begin{IEEEbiography}[{\includegraphics
[width=1in,height=1.25in,clip,
keepaspectratio]{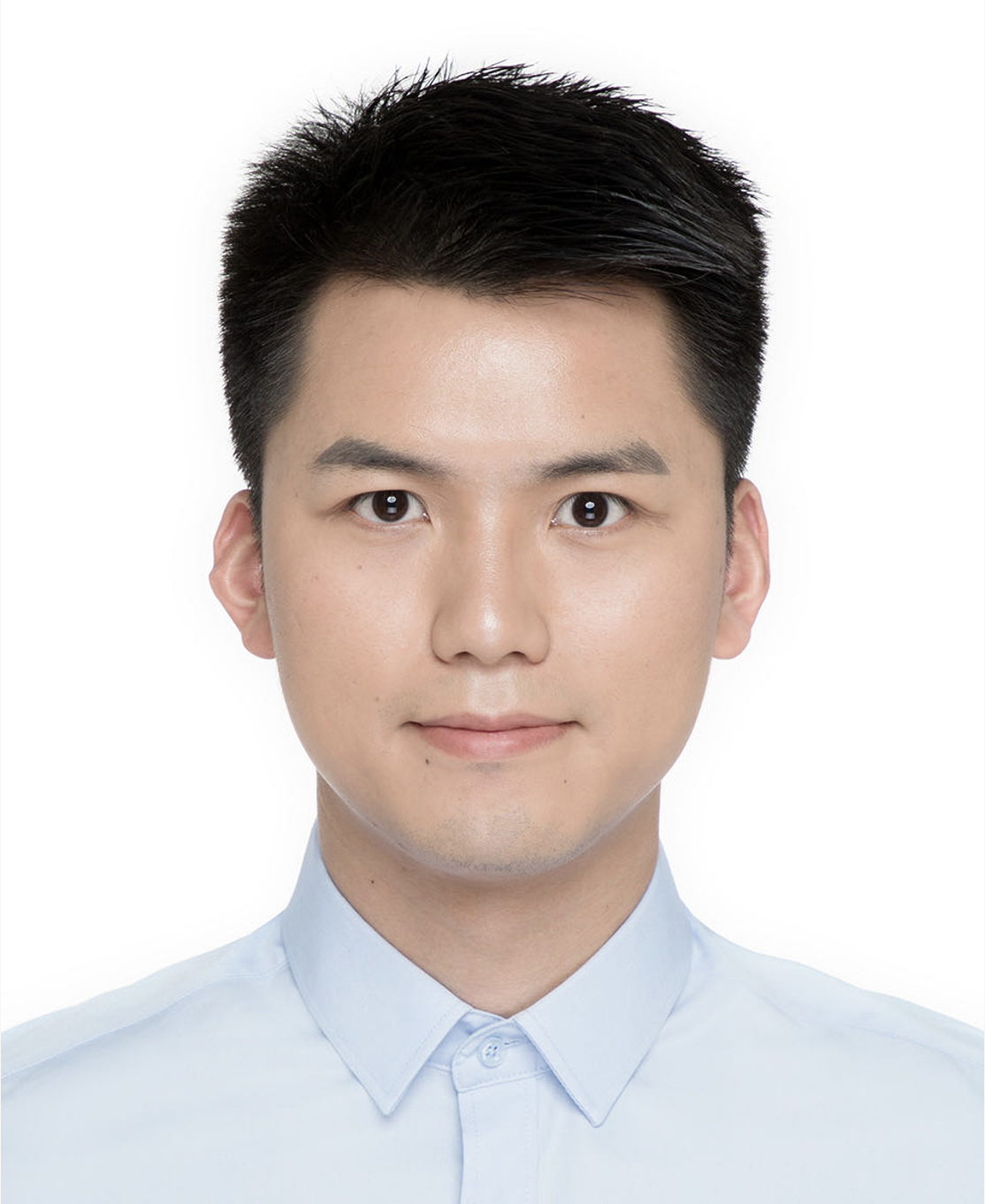}}]
{Lei Li} is a PhD student in the HKU-NLP group at the University of Hong Kong, co-supervised by Prof. Lingpeng Kong and Prof. Qi Liu. He obtained his master's degree from Peking University under the supervision of Prof. Xu Sun and his bachelor's degree from Xidian University. His research centers on multimodal large language models and investigating the fundamental mechanisms underlying large language models. He serves as Area Chair for ACL ARR and has been a reviewer for leading venues including IJCV, NeurIPS, ICML, CVPR, and ICLR.
\end{IEEEbiography}

\begin{IEEEbiography}[{\includegraphics
[width=1in,height=1.25in,clip,
keepaspectratio]{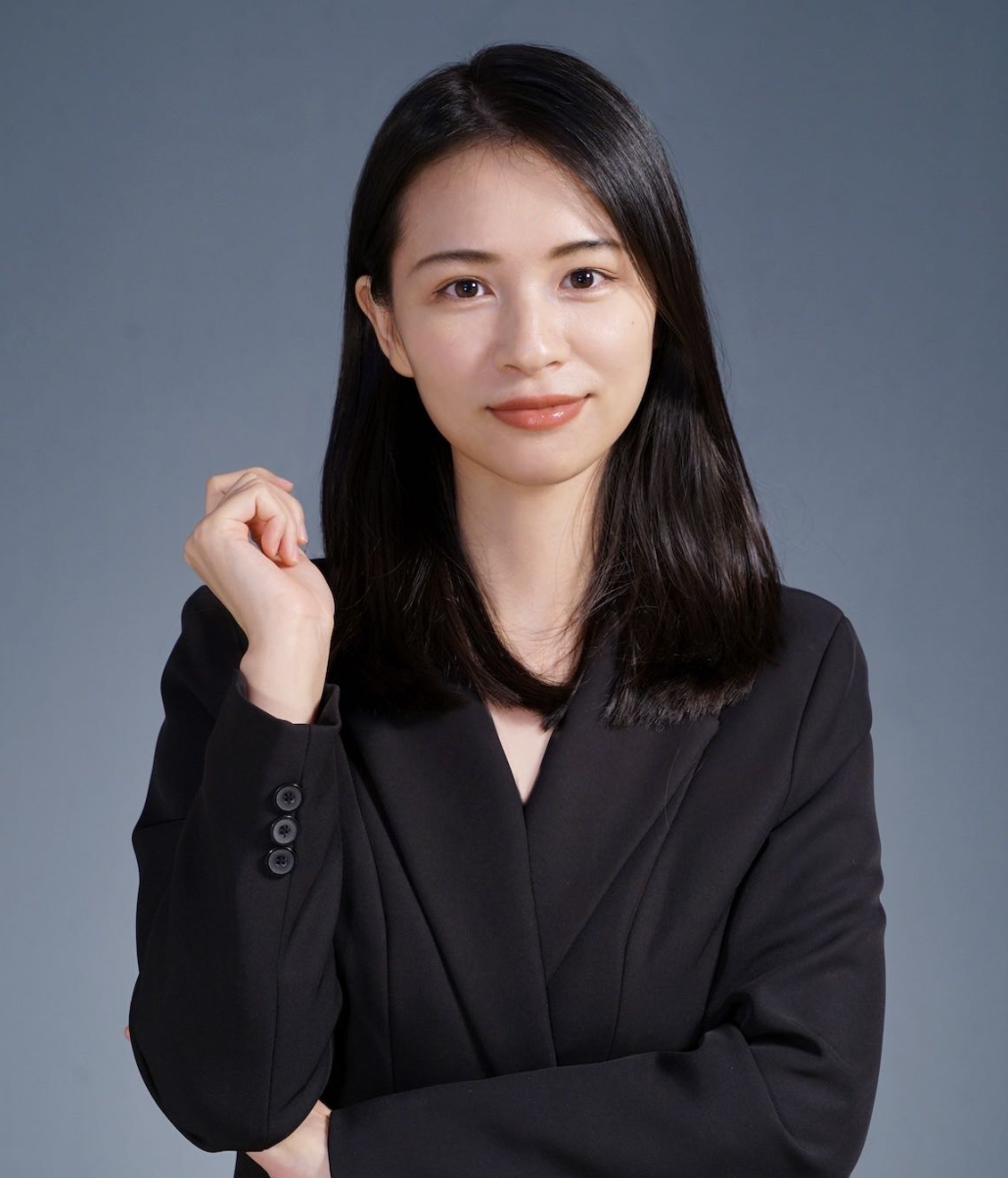}}]
{Lin Qiu} is an assistant professor at the Southern University of Science and Technology. Her research interests include the interplay between computer science and information systems management. She holds a Ph.D. from the National University of Singapore. Her research appears in IEEE Transactions on Knowledge and Data Engineering, Production and Operations Management, Journal of Medical Internet Research, and ACM Transactions on Management Information Systems.
\end{IEEEbiography}

\begin{IEEEbiography}[{\includegraphics
[width=1in,height=1.25in,clip,
keepaspectratio]{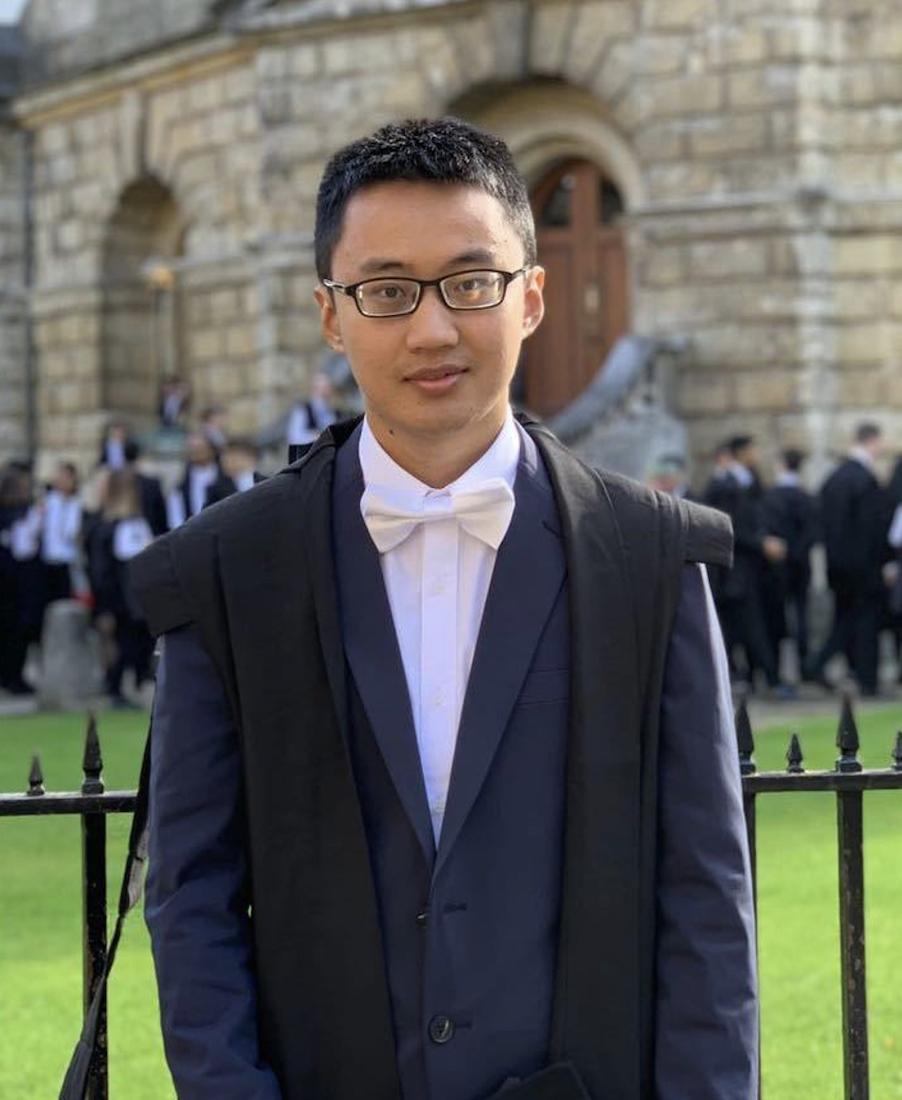}}]
{Qi Liu} is an assistant professor at the Department of Computer Science, the University of Hong Kong, and a cofounder of Reka. He earned his Ph.D. in computer science from the University of Oxford. In the past, he obtained a Master of Science degree from the National University of Singapore, and a Bachelor of Engineering degree from Shandong University. His research interests include natural language processing and machine learning. His research is centered on enabling computers to comprehend human language. He also spent some time at Google DeepMind, Facebook AI Research, and Microsoft Research before and during his PhD study.
\end{IEEEbiography}

\end{document}